\documentclass[journal]{IEEEtran}
\newcommand{\ignore}[1]{}

\usepackage[utf8]{inputenc}
\usepackage[T1]{fontenc}
\usepackage{amsmath,epsfig}
\usepackage{url}
\usepackage[pagebackref=true,breaklinks=true,letterpaper=true,colorlinks,citecolor=green,linkcolor=green,bookmarks=false]{hyperref}
\usepackage{graphicx}
\usepackage{bm}
\usepackage{amssymb}
\usepackage{cite}
\usepackage{array}
\usepackage{color}
\usepackage{booktabs}
\usepackage{tabu}

\usepackage{multicol}
\usepackage{lettrine}
\usepackage{multirow}
\usepackage{picins}

\usepackage{fixltx2e} 

\begin{document}
%
\title{Interactive Feature Embedding for Infrared and Visible Image Fusion}

\author{Fan Zhao, Wenda Zhao, Huchuan Lu
\thanks{

Fan Zhao is with School of Physics and Electronic Technology, Liaoning Normal University, Dalian, 116029, P.R. China. \protect
E-mail: Fan\_Zhao20@163.com.

Wenda Zhao and Huchuan Lu are with the Key Laboratory of Intelligent Control and Optimization for Industrial Equipment of Ministry of Education and School of Information and Communication Engineering, Dalian University of Technology, Dalian, 116024, P.R. China. \protect
E-mail: zhaowenda@dlut.edu.cn,lhchuan@dlut.edu.cn.
}
}


\maketitle

\begin{abstract}
General deep learning-based methods for infrared and visible image fusion rely on the unsupervised mechanism for vital information retention by utilizing elaborately designed loss functions. However, the unsupervised mechanism depends on a well designed loss function, which cannot guarantee that all vital information of source images is sufficiently extracted.
In this work, we propose a novel interactive feature embedding in self-supervised learning  framework for infrared and visible image fusion, attempting to overcome the issue of vital information degradation.
With the help of self-supervised learning framework, hierarchical representations of source images can be efficiently extracted.
In particular, interactive feature embedding models are tactfully designed to build a bridge between the self-supervised learning and infrared and visible image fusion learning, achieving vital information retention.
Qualitative and quantitative evaluations exhibit that the proposed method performs favorably against state-of-the-art methods.
\end{abstract}


\IEEEpeerreviewmaketitle

\section{Introduction}
\lettrine[lines=2]{\textbf{V}}{ISIBLE} image, captured by the reflected light of a scenario, contains abundant texture details. Complementarily, infrared image exhibits strong anti-interference ability (e.g., smoke and night) by capturing the thermal radiation.
However, the detailed structure information is insufficiently in infrared image.
Infrared and visible image fusion aims to construct a high-quality image by integrating the vital information from source images, which is more conducive to
subsequent applications, such as security, monitoring, target tracking, and object recognition ~\cite{Fan2015A,Raghavendra2011Particle,Ulusoy2011New,9404315}.

\begin{figure*}[t]
	\begin{center}
	\begin{tabular}{c@{}c@{}c@{}c}
		\includegraphics[width=0.24\linewidth,height=3.3cm]{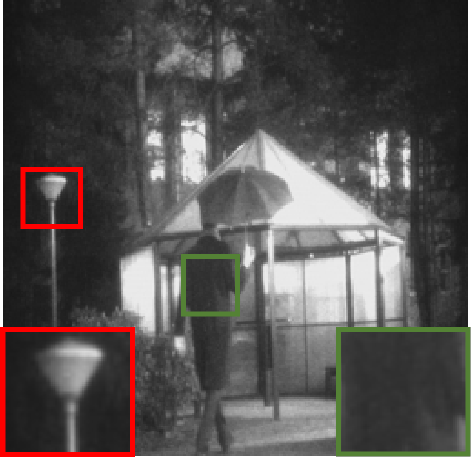} \ &
		\includegraphics[width=0.24\linewidth,height=3.3cm]{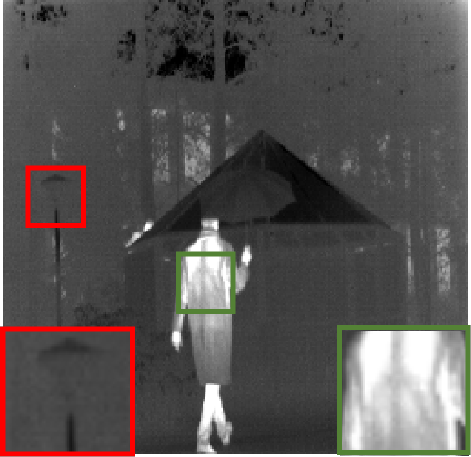} \ &
		\includegraphics[width=0.24\linewidth,height=3.3cm]{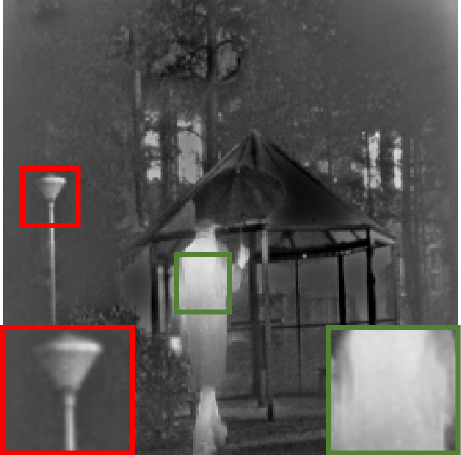} \ &
		\includegraphics[width=0.24\linewidth,height=3.3cm]{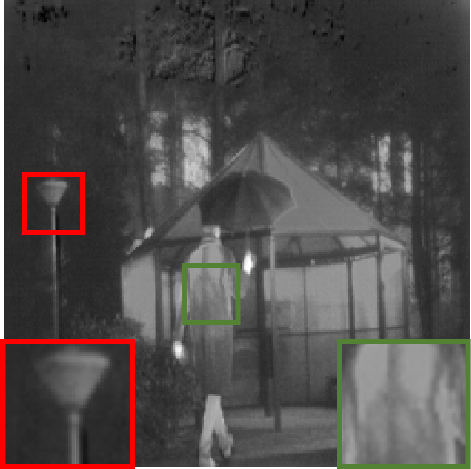} \\
		\small {(a) } \ & \small {(b) } \ & \small {(c) } \ & \small {(d) }\\
		\includegraphics[width=0.24\linewidth,height=3.3cm]{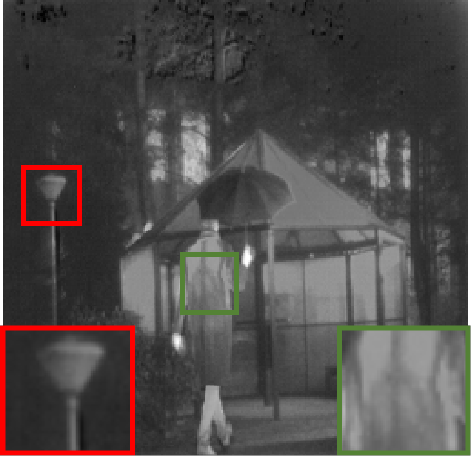} \ &
		\includegraphics[width=0.24\linewidth,height=3.3cm]{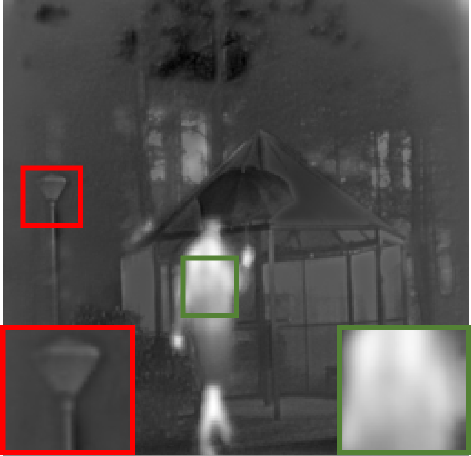} \ &
		\includegraphics[width=0.24\linewidth,height=3.3cm]{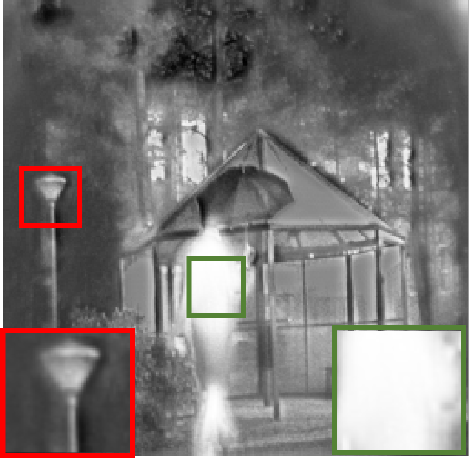} \ &
		\includegraphics[width=0.24\linewidth,height=3.3cm]{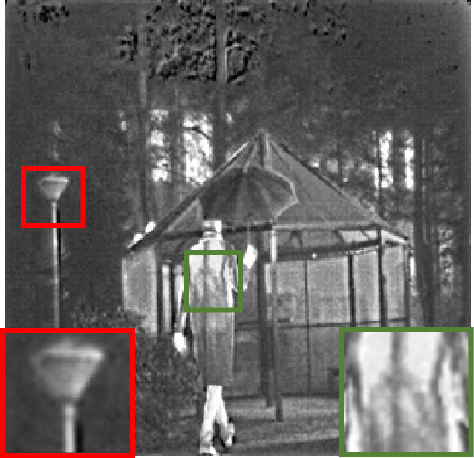} \\
		\small {(e) } \ & \small {(f) } \ & \small {(g) } \ & \small {(h) }\\
	\end{tabular}
    \end{center}
    \vspace{-1mm}
\caption{Vital feature retention in fused images of different methods. (a-h) are visible image, infrared image, GTF, DeepFuse, DenseFuse, FusionGAN, DDcGAN and our IFESNet.}
\label{fig1}
\end{figure*}

Vital feature extraction and fusion are keys associated with infrared and visible image fusion. On the one hand, visible image mainly represents the reflected light information with detailed textures, while infrared image depicts the thermal radiation information with high contrast pixel intensities (as shown in Fig.~\ref{fig1}(a-b). These two types of features have domain discrepancy, which needs special attention.
On the other hand, both visible and thermal images contain some common vital features, such as brightness and object semantics.
Thus, how to comprehensively extract and fuse aforementioned features, including universal features and domain discrepancy features, remains the major stumbling block.

To alleviate the above problems, previous methods can be mainly divided into three categories.
(1) Hand-crafted feature-based methods implement image transform (e.g., multi-scale transform (MST)~\cite{Chen2019,Li2011,JIN2018} and hybrid models~\cite{Liu2015,Zhou2015,Yin2016}) to extract some specific information, such as contrast and textures.
(2) Convolutional neural network (CNN)-based methods (e.g., DenseFuse~\cite{Li2018}, IFCNN~\cite{Zhang2019} and U2Fusion~\cite{9151265}) learn to extract multi-level features, thereby fusing vital and domain discrepancy information.
(3) Adversarial learning-based methods~\cite{Ma2019s,Ma2020,Ma2019ss} intend to fuse thermal radiation information and textural detail information through adversarial training along with designed loss.

Although infrared and visible image fusion has made progress, previous methods generally have the following limitations.
On the one hand, the same image transform or convolution operator can hardly extract comprehensive features.
On the other hand, single-stage feature fusion may not make the fused image retain all vital features of source images.
As shown in Fig.~\ref{fig1}, the intensity information in visible image (e.g., the pavilion) and the texture information in infrared image (e.g., the man) are lost in Fig.~\ref{fig1}(c), the intensity information of source images is not well preserved in Fig.~\ref{fig1}(d) and Fig.~\ref{fig1}(e), and low-contrast dilemmas exist in Fig.~\ref{fig1}(f-g).
Addressing the above problems, we propose self-supervised hierarchical feature extraction and stage-interactive feature fusion framework (IFESNet) for infrared and visible image fusion.

Specifically, we first conceive the self-supervised strategy for jointly source image reconstruction and fusion. As an auxiliary task, image reconstruction task is trained by regarding the source image as ground truth in a self-supervised way.
Therefore, richer and more comprehensive features of source images can be learned.
Compared with unsupervised mechanism, our self-supervised strategy can capture more informative representations.
Furthermore, we propose an interactive feature embedding
model (IFEM) to build a bridge between the
self-supervised learning and infrared and visible image fusion learning, achieving vital
information retention.
IFEM is formulated by interacting hierarchical representations between fusion and reconstruction tasks.
Specifically, the interaction process is recursively conducted in the corresponding hierarchical layers between reconstruction and fusion tasks. Note that the hierarchical representation interaction process is bi-directional.
Therefore, we utilize the internal relationship between different tasks to efficiently extract and fuse these feature representations, improving the performance of fusion task.

As illustrated in Fig.~\ref{fig1}(h), our IFESNet can retain larger amount of thermal radiation information (e.g., person) from Fig.~\ref{fig1}(b) and textural detail information (e.g., pavilions and trees) from Fig.~\ref{fig1}(a). In addition, as shown in the locally enlarged areas, our fusion result can also effectively preserve the texture and edge information of Fig.~\ref{fig1}(b) (e.g., the dorsal area of the person in green box), and high-contrast intensity information of Fig.~\ref{fig1}(a) (e.g., street lamp in red box). On the contrary, GTF, FusionGAN and DDcGAN lose partial texture details in infrared image, while DeepFuse and DenseFuse lose some contrast information in visible image. Concretely, our contributions are:

\begin{itemize} \item We make the first attempt to develop self-supervised strategy to solve the vital information missing dilemmas in infrared and visible image fusion.
In contrast with the most widely used non-adversarial and adversarial fusion methods,
our framework is simple yet effective, better improving the performance of fused image.

\item
Interactive feature embedding
model is designed across fusion and reconstruction tasks, gradually extracting vital information representations and promoting fusion task.

\item
Compared with the state of the arts, our proposal can retain more vital features, including universal features and domain discrepancy features.
\end{itemize}

The remainder of the paper is organized as follows. Section II introduces the related work. Section III presents our proposed framework. In Section IV, qualitatively and quantitatively results are evaluated and analyzed, and ablation studies are conducted. Finally, Section V concludes this paper.

\begin{figure*}[t]
	\begin{center}
		\begin{tabular}{c@{}c}
		\includegraphics[width=0.44\linewidth,height=2.8cm]{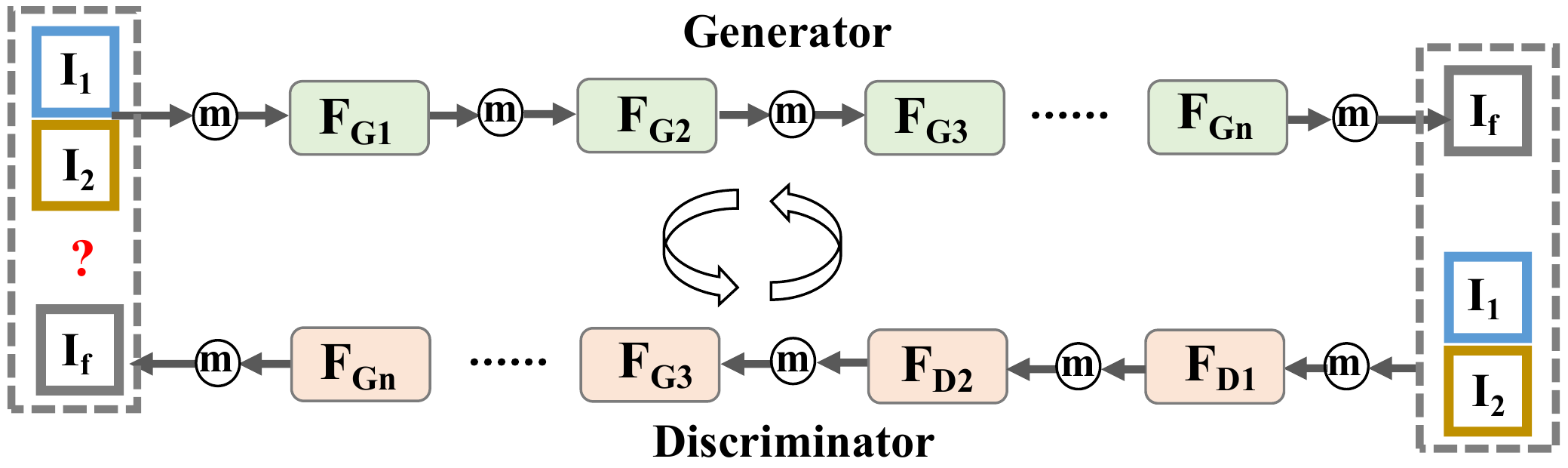} \ &
		\includegraphics[width=0.46\linewidth,height=2.8cm]{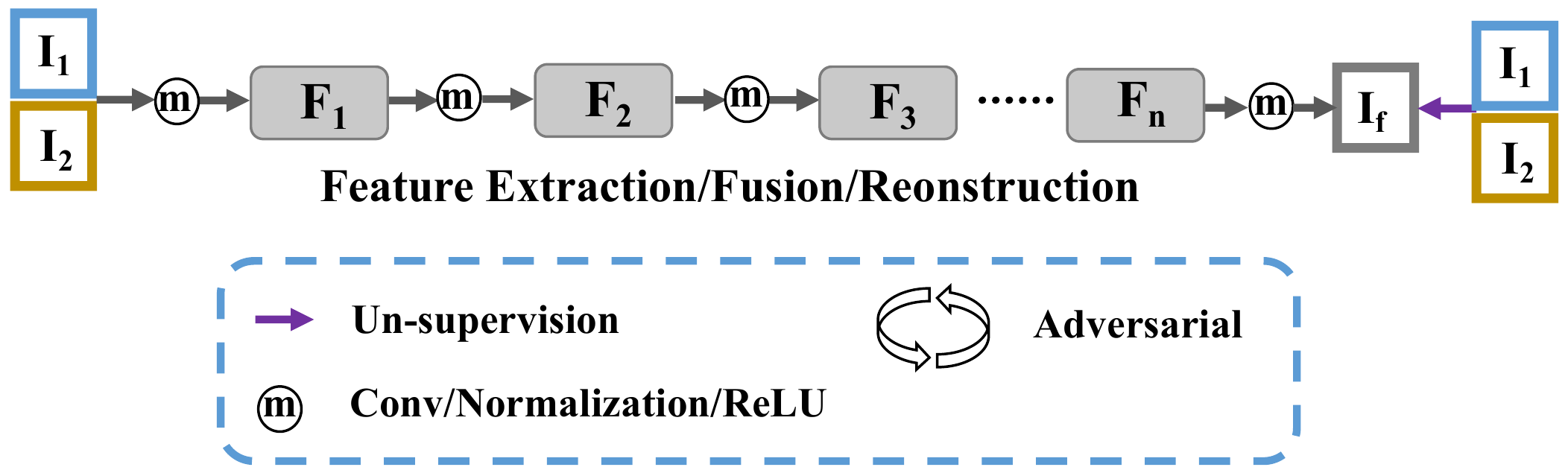} \\
		\small {(a) } \ & \small {(b) }\\
		\end{tabular}
    \end{center}
    \vspace{-1mm}
	\caption{ Pipeline of varied types of fusion methods. (a-b) correspond to the adversarial fusion framework and non-adversarial fusion framework. The colorful box ($F_{G}$, $F_{D}$ and $F_{n}$) represent hierarchical features. The arrows between blocks indicate the feature stream. $I_{1}$, $I_{2}$ are source images. $I_{r1}$, $I_{r2}$ are reconstructed results of source images, and $I_{f}$ represents the final fusion result.}
	\label{fig2}
\end{figure*}

\section{Related Work}
\subsection{Hand-crafted Feature-based Methods}
Varied attempts based on hand-crafted feature extraction have been proposed. For instance, multi-scale transform~\cite{Chen2019,Li2011,JIN2018}, sparse representation (SR)~\cite{Zhang2014,Wang2014,Ma2019,Zhang2018,Lu2014}, subspace~\cite{Zhizhong2016Infrared,Mitianoudis,Kong2014}, hybrid models~\cite{Liu2015,Zhou2015,Yin2016} and gradient~\cite{zhao2016gradient} are widely used. For improving feature extraction ability, Laplace pyramid~\cite{Chen2019}, contourlet~\cite{Li2011}, gradient~\cite{Ma2016Infrared}, SR~\cite{Lu2014}, independent component analysis (ICA)~\cite{Mitianoudis}, principal component analysis (PCA)~\cite{Zhizhong2016Infrared}, and nonnegative matrix factorization (NMF)~\cite{Kong2014}-based methods are developed, attempting to transform the source images with the same operation. As illustrated in Fig.~\ref{fig1}, visible image mainly represents the reflected light information, while infrared image depicts the thermal radiation information. Inherently, these two types of features are specific to source image with domain discrepancy, which can hardly be represented in the same manner. In addition, hand-crafted features can not comprehensively represent all features of source images, resulting in limited performance.

\subsection{Non-adversarial Fusion Methods}
Deep learning models ~\cite{zhao2021self,9069930,Liu2017,Li2018s,zhao2018multi,Li2018,9151265,Zhang2020,8295275,9106801,zhao2021depth} for image fusion have been put forward, whereas feature extraction and fusion still remain an active topic. Liu $et~al.$~\cite{Liu2017} firstly utilize a siamese convolutional network for weight map generation. To avoid the loss of vital features, similarity-based fusion strategy is adopted to adjust the decomposed coefficients. Li $et~al.$~\cite{Li2018s} decompose infrared and visible images into base layers and detail content layers, and then use CNN for feature extraction of detail content layers. To boost feature extraction performance, DenseFuse~\cite{Li2018} adopts dense block to construct feature maps. Xu $et~al.$~\cite{9151265} propose U2Fusion, where the similarity between fusion result and source images is constraint for improving the vital information retention in an unsupervised way. For further enhancing the texture and intensity information, Zhang $et~al.$~\cite{Zhang2020} develop proportional maintenance of gradient and intensity (PMGI) network, in which path-wise transfer block is designed to avoid the vital information loss.

Those fusion frameworks, in spirit, can be roughly summarized as Fig.~\ref{fig2}(b). As a whole, they focus on extracting and fusing vital features via unsupervised strategy, while loss function is designed to constrain the fusion results to contain the desired features. However, as illustrated in Fig.~\ref{fig1}, it can hardly design a comprehensive and adaptive loss function covering all vital features. In other words, unsupervised mechanism is not capable enough to extract and fuse all vital features of source images.

\subsection{Adversarial Fusion Methods}
Adversarial fusion methods aim to formulate the fusion task as an adversarial learning, in which the thermal radiation information is fused while the visible texture information is retained. Ma $et~al.$~\cite{Ma2019s} develop FusionGAN, which generates the fusion results with infrared intensities and dominant visible gradients. On this basis, DDcGAN~\cite{Ma2020} adopts a generator with two discriminators to enhance the edge information of thermal targets with the specially designed content loss. Later, Ma $et~al.$~\cite{Ma2019ss} design a variant based on detail preserving adversarial learning, where detail loss and target edge-enhancement loss are designed based on FusionGAN. This approach can greatly retain the vital features of source images compared with FusionGAN.

The overview of adversarial fusion methods can be roughly summarized as Fig.~\ref{fig2}(a). A generator is adopted to produce a fused image $I_f$ with major features of source images, and then a discriminator is used to distinguish $I_f$ from source images. Hence, fusion task is achieved by this adversarial learning mechanism. However, adversarial process is difficult to optimize, which easily leads to distorted results. In addition, the feature diversity of source images, not only thermal radiation and gradient variation, but also the other common features, brings great difficulty for comprehensively covering all vital features. Consequently, any features (e.g., textures in infrared image or intensities in visible image) ignored by the loss function, will not be retained in the fusion results (as shown in Fig.~\ref{fig1}(f-g)). Similarly, this can also be attributed to the limitation of unsupervised mechanism.

\begin{figure*}[t]
	\begin{center}
	\includegraphics[width=1\linewidth,height=5.3cm]{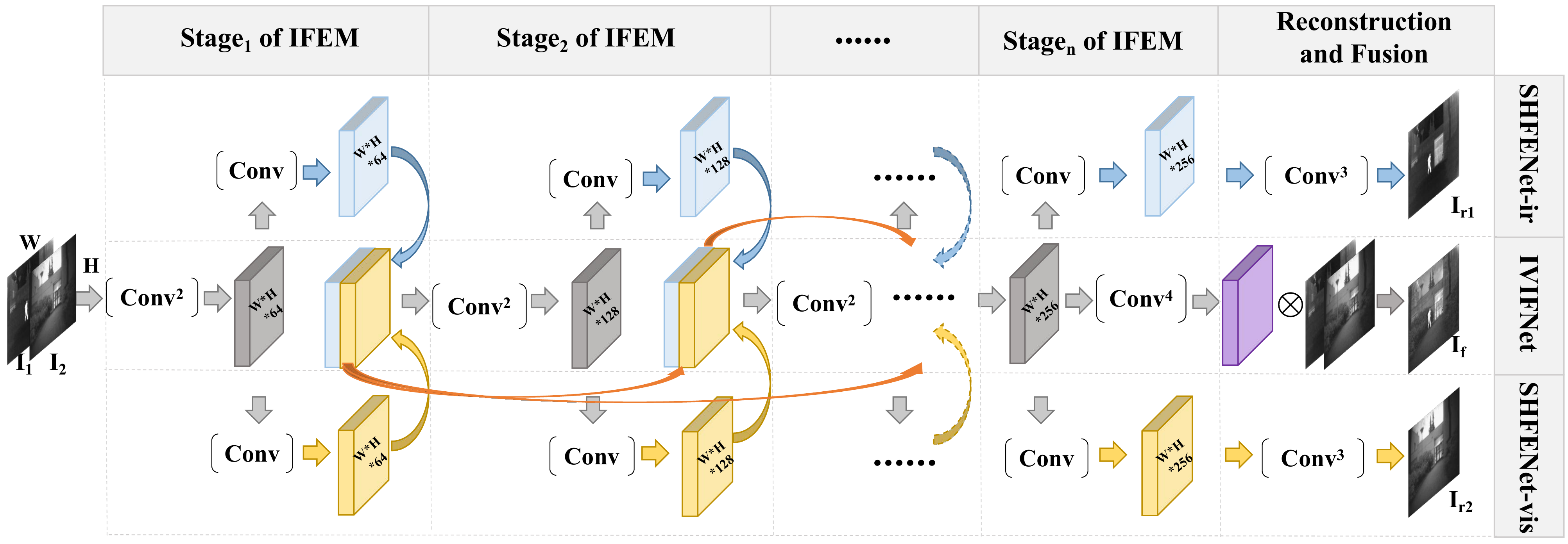}
    \end{center}
    \vspace{-1mm}
	\caption{The detailed architecture of interactive feature
embedding in self-supervised learning network (IFESNet). In the horizontal direction, IFESNet is consisted by self-supervised hierarchical feature extraction network of infrared image (SHFENet-ir), self-supervised hierarchical feature extraction network of visible image (SHFENet-vis), and infrared and visible image fusion network (IVIFNet), where self-supervised hierarchical features are used for jointly reconstruction and fusion. In the vertical direction, IFESNet is consisted by multi stages of interactive feature
embedding models (IFEM), which conducts hierarchical feature interaction stage-interactive between IVIFNet and SHFENet to gradually fuse vital information, thereby avoiding vital feature loss in fusion result. $\otimes$ indicates channel-wise multiplication. $Conv^{n}$ denotes $n$ times convolution operations. Note that only convolution layers are used for feature extraction.}
	\label{fig3}
\end{figure*}

\section{Proposed Method}

\subsection{Motivation}\label{3.1}
We are committed to alleviate vital feature loss dilemma. Infrared image and visible image are characterized by thermal radiation and visible gradient information, respectively.
Their domain discrepancy needs special attention. On the other hand, they contain common attribute features, such as gradient variation, intensity, contrast and saturability.
Majority fusion methods adopt the same operator for feature extraction, resulting in limited performance for dealing with domain discrepancy features. Most importantly, existing fusion methods involve unsupervised strategy with elaborate loss functions for vital feature fusion. Such mechanism is not adequately for retaining all vital information. Since it is infeasible to design a comprehensive and adaptive loss function that covers all vital features. Ignoring any information (e.g., texture in infrared or intensity in visible) will result in vital feature missing (as shown in Fig.~\ref{fig1}).

In this section, we develop a novel interactive feature embedding in self-supervised learning network (IFESNet) for infrared and visible image fusion. Different from the widely used unsupervised mechanism, we attempt to develop self-supervised strategy in cooperation with stage-interactive feature embedding learning to solve the issue of vital information missing. The pipeline is illustrated in Fig.~\ref{fig3}. Several concepts have been considered to conceive such architecture, including 1) self-supervised hierarchical feature extraction for jointly reconstruction and fusion tasks in Section III.B, and 2) stage-interactive feature embedding learning in Section III.C.
Please note that since pooling operation will reduce spatial resolution of features, our IFESNet consists of convolution layers to retain spatial details of the fused image effectively.

\subsection{Self-supervised Hierarchical Feature Extraction}
By self-supervised mechanism, we aim to achieve hierarchical feature extraction containing more informative representations, thereby boosting fusion performance.
As shown in Fig.~\ref{fig3}, IFESNet includes SHFENet-ir (self-supervised hierarchical feature extraction network of infrared image), SHFENet-vis (self-supervised hierarchical feature extraction network of visible image), and IVIFNet (infrared and visible image fusion network). In the following section, we will detail the self-supervised strategy for jointly reconstruction and fusion tasks.

\textbf{Self-supervised Feature Extraction.} Vital feature extraction is the premise of boosting fusion performance. Thus, we aim to extract hierarchical features $F_{n}^{\prime}$ and $F_{n}^{\prime\prime}$, which contain comprehensive features of infrared image $I_{1}$ and visible image $I_{2}$. Essentially, this is achieved through reconstructing images $I_{r1}$ and $I_{r2}$ using the source image as ground truth in a self-supervised way. Since hierarchical features $F_{n}^{\prime},F_{n}^{\prime\prime}$ can be reconstructed back to source images, it in turn ensures that the corresponding hierarchical layers are competent to extract vital features of source images.

Take the SHFENet-ir as an example (see Fig.~\ref{fig3}). From the perspective of information flow, the SHFENet-ir is constructed by hierarchical feature interaction process between each layer of SHFENet-ir and IVIFNet. In particular, the hierarchical features $F_{n}^{\prime}$ and $F_{n}^{\prime\prime}$ are totally obtained from the hierarchical feature $F_{n}$ of IVIFNet, which can be formulated as
\begin{equation}
F_{n}^{\prime}, F_{n}^{\prime\prime} =C(\underbrace{C^{2}(Cat(F_{n-1}^{\prime}, F_{n-1}^{\prime\prime}))}_{F_{n}}),
\end{equation}
where $F_{n}^{\prime}$ and $F_{n}^{\prime\prime}$ denote the hierarchical features of layer $n$ in SHFENet-ir and SHFENet-vis, respectively. $F_{n}$ is hierarchical feature of IVIFNet. $C$ and $C^{2}$ represent conducting convolution operation one and twice, respectively. $Cat$ denotes the concatenation operation.
For the last layer of SHFENet, the outputs are the reconstructed results, which can be formulated as
\begin{equation}
I_{r1}= C^{3}(F_{n}^{\prime}),
I_{r2}= C^{3}(F_{n}^{\prime\prime}),
\end{equation}
where $I_{r1}$ and $I_{r2}$ denote reconstructed results of infrared and visible images, respectively. $C^{3}$ represents conducting convolution operation three times.

The self-supervised strategy ensures that hierarchical features $F_{n}^{\prime}, F_{n}^{\prime\prime}$ contain the main features of source image $I_{1},I_{2}$ during reconstruction task. It is worth to note that the hierarchical features $F_{n}^{\prime},F_{n}^{\prime\prime}$ of reconstruction task are totally obtained from the corresponding hierarchical features $F_{n}$ of fusion task. Thus, $F_{n}^{\prime}, F_{n}^{\prime\prime}$ in turn constrain the hierarchical feature $F_{n}$ of IVIFNet to possess the main features of $I_{1},I_{2}$. In other words, the self-supervised strategy promotes the fusion task.
Specifically, SHFENet is constructed by six convolution layers with 3*3 kernels and 64, 128, 256, 128, 64 and 1 channels respectively. Note that down-sampling and up-sampling structures are not adopted in SHFENet, which can avoid the loss of effective information.

\textbf{Hierarchical Feature Fusion.} In IVIFNet, we aim to generate the fusion result utilizing hierarchical features obtained by self-supervised mechanism. Considering the extracted hierarchical features $F_{n}^{\prime},F_{n}^{\prime\prime}$ cover sufficient information of source images, it is potential for encouraging fusion task. Thus, how to utilize these vital features for fusion task remains a problem to be solved.

To achieve this, the hierarchical feature interaction process between reconstruction and fusion task is designed to gradually promote the fusion network. As shown in Fig.~\ref{fig3}, we first concatenate the source images $I_{1}$ and $I_{2}$, and then the hierarchical feature $F_{1}$ can be obtained by conducting convolution operation $Conv$ twice. Based on the idea of merging and then separating features \cite{gao2019nddr}, we fuse the layer-wise hierarchical features $F_{n-1}^{\prime}$ and $F_{n-1}^{\prime\prime}$ from SHFENet, and the hierarchical feature $F_{n}$ of IVIFNet can be formulated as
\begin{equation}
F_{n}=C^{2}(Cat(\underbrace{C(F_{n-1})}_{F_{n-1}^{\prime}}, \underbrace{C(F_{n-1})}_{F_{n-1}^{\prime\prime}}))).
\end{equation}

At this point, the hierarchical feature $F_{n}$ of IVIFNet is derived from hierarchical features $F_{n-1}^{\prime},F_{n-1}^{\prime\prime}$, which heuristically shares low-, mid- and high-level features for fusion. Thus, the extracted hierarchical features $F_{n}^{\prime},F_{n}^{\prime\prime}$ via self-supervised strategy can be fully utilized for fusion task, which in turn avoids the vital features missing in fusion result.
Specifically, IVIFNet is constructed by ten convolution layers with 3*3 kernels and 64, 64, 128, 128, 256, 256, 256, 128, 64 and 1 channels respectively. The outputs of the last convolution layer of IVIFNet are weight maps for infrared and visible images. Thus, the fusion result can be generated by using channel-wise multiplication with source images $I_{1},I_{2}$, which can be written as
\begin{equation}
I_f =\sum_{i=1}^2\underbrace{C^{4}(F_{n})}_{W_i} \otimes I_i,
\end{equation}
where $I_f$ denotes the fusion result. $W_i$ is the $i$th weight map, which is calculated by conducting four convolutions on $F_n$. $I_i$ represents $i$th source image.

\subsection{Stage-interactive Feature Embedding Model}\label{3.2.2}
As shown in Fig.~\ref{fig3}, vertically, our framework is consisted by multi-stage interactive feature embedding models (IFEMs). IFEMs conduct hierarchical feature interaction between SHFENet and IVIFNet. This allows to jointly learn correlative representations for alleviating vital feature missing in fusion result. We argue that the layers of SHFENet and IVIFNet can be treated as different feature descriptors, and features learned from different tasks can be treated as different representations of source images. Thus, feature representations related to reconstruction can provide additional vital features for fusion. Specifically, hierarchical feature interaction is conducted as a bridge between reconstruction and fusion task, which can utilize the internal relationship between different tasks to improve the feature representations, thereby boosting the performance of fusion task.
The $n$-stage IFEM for hierarchical feature interaction between $F_{n}$ and $F_{n-1}^{\prime},F_{n}^{\prime},F_{n-1}^{\prime\prime},F_{n}^{\prime\prime}$ can be expressed as
\begin{equation}
F_{n}^{\prime},F_{n}^{\prime\prime} = Bi(F_{n}), F_{n} = INT(F_{n-1}^{\prime},F_{n-1}^{\prime \prime}),
\end{equation}
where $Bi$ denotes the hierarchical feature delivering process from IVIFNet to SHFENet, and $INT$ represents the hierarchical feature delivering process from SHFENet to IVIFNet, which are equivalent to formulas (1) and (3), respectively. $n$ denotes the stage number.
Whereas the interaction process is bi-directional, $F_{n}$ and $F_{n}^{\prime},F_{n}^{\prime\prime}$ can be represented by each other.

To be specific, the $INT$ process is designed to ensure that vital hierarchical features $F_{n}^{\prime},F_{n}^{\prime\prime}$ are concatenated and shared to the corresponding hierarchical layer of fusion network, thereby promoting the fusion task. The stage-interactive $INT$ process can greatly reduce the loss of intermediate information by leveraging all the hierarchical features for fusion. The $Bi$ process aims to deliver the fused feature $F_{n}$ to the reconstruction task, which in turn ensures that fused hierarchical feature $F_{n}$ contains important information of source images. Therefore, stage-interactive feature embedding learning for hierarchical feature interaction between the reconstruction and fusion tasks can improve fusion performance.
\begin{table*}
	\small
	\renewcommand{\arraystretch}{1.5}
\setlength\tabcolsep{9pt}%
	\begin{center}
		\caption{Average quality metrics of different methods on INFV-41 dataset. The best result is shown in \textcolor[rgb]{1,0,0}{\textbf{red}}, and second result is shown in \textcolor[rgb]{0,0,1}{\textbf{blue}}.}
		\label{tab:precision1}
		\vspace{0.1cm}
		\begin{tabular}{ccccccccc}
			\toprule[1pt]
			& GTF & DTCWT &  DeepFuse & DenseFuse & IFCNN & U2Fusion & FusionGAN & Ours\\
			\hline\hline
			AG &{4.888}  &{7.154}  &{6.153} &{6.161} &\textcolor[rgb]{0,0,1}{\textbf{8.752}} &{7.458} &{3.662} &\textcolor[rgb]{1,0,0}{\textbf{10.18}}\\	
			\cline{1-9}
			EN  &{6.669}  &{6.610}  &{6.885} &{6.872}  &{6.847} &\textcolor[rgb]{0,0,1}{\textbf{6.920}} &{6.588} &\textcolor[rgb]{1,0,0}{\textbf{6.923}}\\	
			\cline{1-9}
			MI &{13.34} &{13.22}  &{13.77}&{13.75}  &{13.69} &\textcolor[rgb]{0,0,1}{\textbf{13.84}} &{13.18} &\textcolor[rgb]{1,0,0}{\textbf{13.85}}\\	
			\cline{1-9}
			GLD &{8.562} &{12.49} &{10.85}  &{10.85} &\textcolor[rgb]{0,0,1}{\textbf{15.28}} &{13.17} &{6.638} &\textcolor[rgb]{1,0,0}{\textbf{17.96}}\\	
			\cline{1-9}
			SF &{.0415}  &{.0603}    &{.0488}  &{.0497} &\textcolor[rgb]{0,0,1}{\textbf{.0678}} &{.0597} &{.0294} &\textcolor[rgb]{1,0,0}{\textbf{.0787}}\\	
			\cline{1-9}
			VIFF &{.2205} &{.3834}  &{.6080} &{.5988} &{.4774} &\textcolor[rgb]{1,0,0}{\textbf{.6955}} &{.2923} &\textcolor[rgb]{0,0,1}{\textbf{.6414}}\\	
			\toprule[1pt]
		\end{tabular}
	\end{center}
	\vspace{-1mm}
\end{table*}

\begin{table*}
	\small
	\renewcommand{\arraystretch}{1.5}
\setlength\tabcolsep{9pt}%
	\begin{center}
		\caption{Average quality metrics of different methods on INFV-20 dataset. The best result is shown in \textcolor[rgb]{1,0,0}{\textbf{red}}, and second result is shown in \textcolor[rgb]{0,0,1}{\textbf{blue}}.}
		\label{tab:precision1}
		\vspace{0.1cm}
		\begin{tabular}{ccccccccc}
			\toprule[1pt]
			& GTF & DTCWT  & DeepFuse & DenseFuse & IFCNN & U2Fusion & FusionGAN & Ours\\
			\hline\hline
			AG  &{4.772}  &{5.920}  &{5.199} &{5.161} &\textcolor[rgb]{0,0,1}{\textbf{6.734}} &{6.699} &{3.293} &\textcolor[rgb]{1,0,0}{\textbf{9.924}}\\	
			\cline{1-9}
			EN  &{6.611}  &{6.414}  &{6.732} &\textcolor[rgb]{1,0,0}{\textbf{6.899}} &{6.638} &{6.783} &{6.449} &\textcolor[rgb]{0,0,1}{\textbf{6.787}}\\	
			\cline{1-9}
			MI  &{13.22}  &{12.83}  &{13.46} &\textcolor[rgb]{1,0,0}{\textbf{13.80}} &{13.28} &{13.57} &{12.90} &\textcolor[rgb]{0,0,1}{\textbf{13.58}}\\	
			\cline{1-9}
			GLD &{8.177}  &{10.11}   &{9.011}  &{8.876} &\textcolor[rgb]{0,0,1}{\textbf{11.56}} &{11.54} &{5.759} &\textcolor[rgb]{1,0,0}{\textbf{16.99}}\\	
			\cline{1-9}
			SF  &{.0398}  &{.0490}   &{.0397}  &{.0431} &\textcolor[rgb]{0,0,1}{\textbf{.0526}} &{.0505} &{.0266} &\textcolor[rgb]{1,0,0}{\textbf{.0770}}\\	
			\cline{1-9}
			VIFF &{.1931}  &{.3484}   &{.6084}  &{.4441} &{.4924} &\textcolor[rgb]{0,0,1}{\textbf{.6496}} &{.2504} &\textcolor[rgb]{1,0,0}{\textbf{.7011}}\\	
			\toprule[1pt]
		\end{tabular}
	\end{center}
	\vspace{-1mm}
\end{table*}

\subsection{Model Training}\label{3.2.3}

We aim to design a loss function to achieve vital feature extraction and fusion via interactive feature embedding in self-supervised learning framework.
Specifically, we jointly train the reconstruction and fusion tasks. Thus, the designed loss can be expressed as
\begin{equation}
L=L_I+L_V+L_F+L_M,
\end{equation}
where $L_I$ and $L_V$ are self-supervised reconstruction loss functions for SHFENet-ir and SHFENet-vis, respectively. $L_F$ denotes the loss function for IVIFNet. $L_M$ stands for weight map constraint.

Previous consistent loss functions (e.g., energy-based contrast constraint~\cite{zhao2020learning} and perceptual constraint~\cite{lu2019unsupervised}) generally fuse some specific information, such as luminance and edge.
In contrast, we introduce a self-supervision constraint through designing a image
reconstruction task, which is trained by regarding the source image
as ground truth.
Thus, more comprehensive features of source images can be learned.
Here, we adopt the standard mean-square error (MSE) as loss function for self-supervised hierarchical feature extraction network training.
\begin{equation}
L_I=MSE(I_1,I_{r1}),
L_V=MSE(I_2,I_{r2}),
\end{equation}
where $I_1$ and $I_2$ represent the visible image and infrared image, $I_{r1}$ and $I_{r2}$ are the reconstructed results.
The above loss functions ensure that hierarchical layers of reconstruction network own the ability of extracting vital features of the source images.

For further fusing the vital features, we adopt a loss function based on structural similarity index metric (SSIM)\cite{deepfuse,2017Robust,2020Fast} for IVIFNet. Specifically, input images ${I_n}={I_n\vert n=1,2}$ are represented by the components of contrast $C$, structure $S$ and luminance $l$ in SSIM framework:
\begin{equation}
I_n=C_n \ast S_n + l_n.
\end{equation}
Contrast $C_n$ and structure $S_n$ are:
\begin{equation}
C_n=\Vert I_n-\mu_{I_n}\Vert,S_n=\frac {I_n-\mu_{I_n}}{\Vert I_n-\mu_{I_n}\Vert},
\end{equation}
where $\mu_{I_n}$ is the average value of $I_n$.
For an expected result $\overline{I}=\xi \overline{C} \ast \overline{S}$, it should contain high contrast as well as the main structure of the source images. Thus, the corresponding contrast $\overline{C}$, structure $\overline{S}$ can be expressed as
\begin{equation}
\overline{C}=\underbrace{maxC_n}_{n=1,2},\overline{S}=\frac {\sum_{n=1}^{2}S_n}{\Vert \sum_{n=1}^{2}S_n\Vert}.
\end{equation}
Finally, the SSIM between the fusion result $I_f$ and the expected result $\overline{I}$ can be calculated by the following function:
\begin{equation}
SSIM=\frac {2\sigma_{\overline{I}I_f}+C}{\sigma^{2}_{\overline{I}}+\sigma^{2}_{I_f}+C},
\end{equation}
where $\sigma^{2}_{\overline{I}}$ and $\sigma^{2}_{I_f}$ represent the variance of $\overline{I}$ and $I_f$. $\sigma_{\overline{I}I_f}$ is the covariance of $\overline{I}$ and $I_f$.
Thus, the loss function for IVIFNet is calculated as
\begin{equation}
L_F=1-SSIM.
\end{equation}

Weight map constraint $L_M$ aims to adjust map values, thereby improving fusion performance. $L_M$ can be written as
\begin{equation}
L_M=|\tau-W_1-W_2|,
\end{equation}
where $\tau$ is a hyperparameter.

\begin{figure*}[t]
\begin{center}
	\begin{tabular}{c@{}c@{}c@{}c@{}c}
		\includegraphics[width=0.19\linewidth,height=2.9cm]{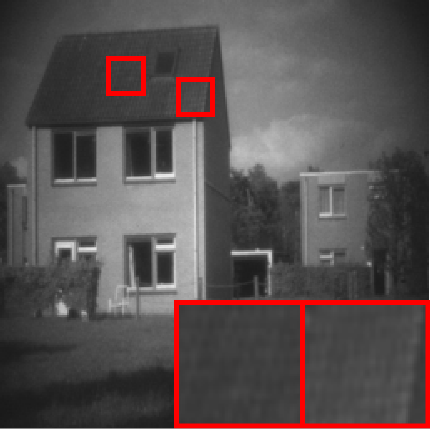} \ &
		\includegraphics[width=0.19\linewidth,height=2.9cm]{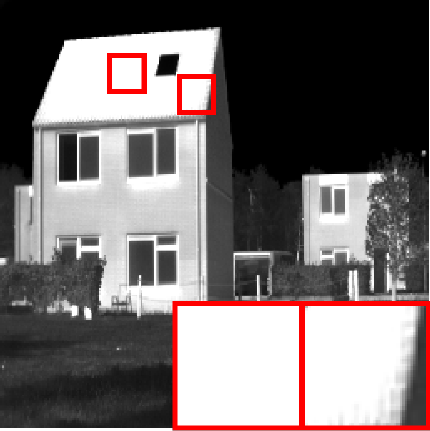} \ &
		\includegraphics[width=0.19\linewidth,height=2.9cm]{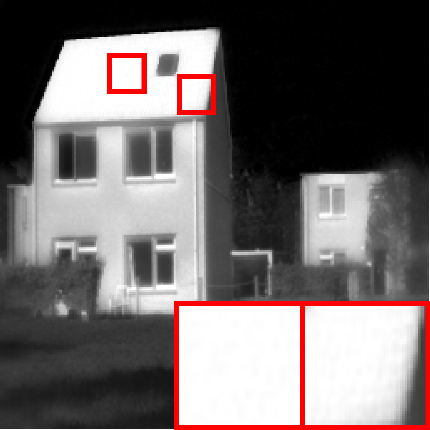} \ &
		\includegraphics[width=0.19\linewidth,height=2.9cm]{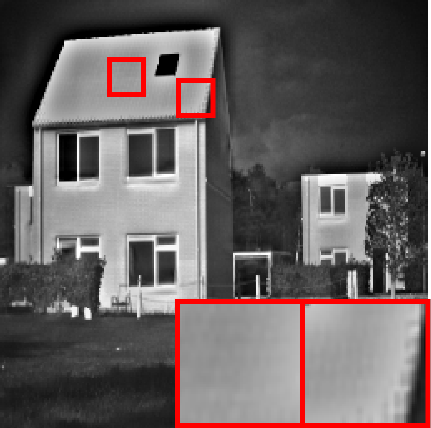}  \ &	
		\includegraphics[width=0.19\linewidth,height=2.9cm]{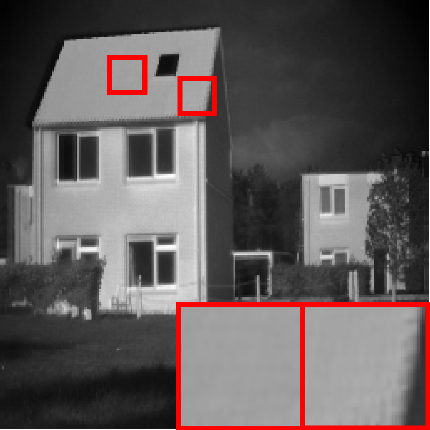} \\
        \small {(a) } \ & \small {(b) } \ & \small {(c) } \ & \small {(d) }\ & \small {(e) }\\
        \includegraphics[width=0.19\linewidth,height=2.9cm]{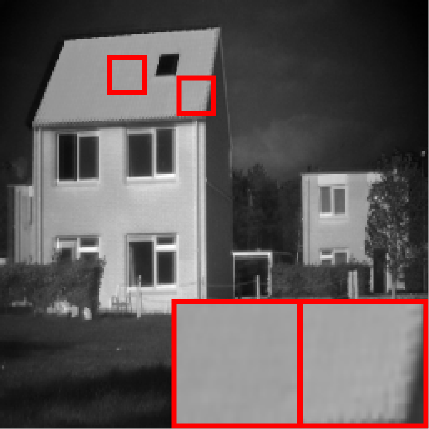} \ &		
        \includegraphics[width=0.19\linewidth,height=2.9cm]{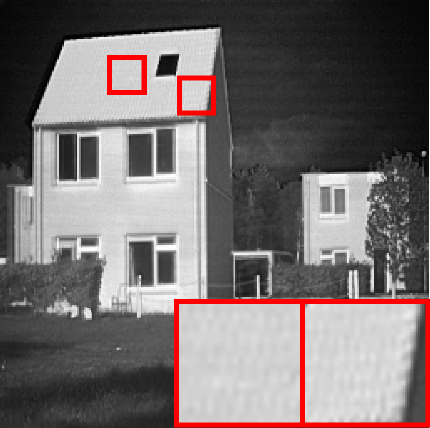} \ &
		\includegraphics[width=0.19\linewidth,height=2.9cm]{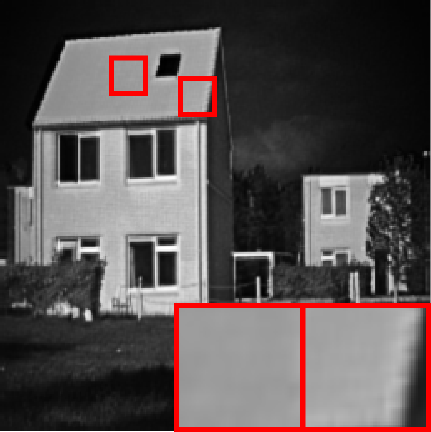} \ &
		\includegraphics[width=0.19\linewidth,height=2.9cm]{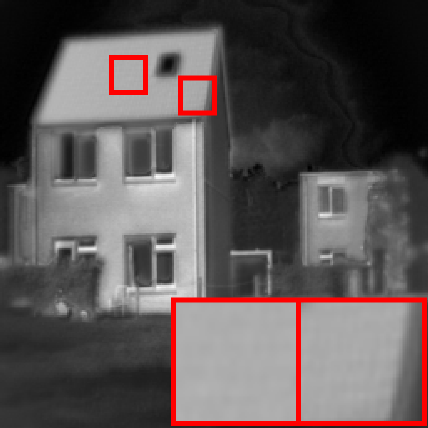}\ &
		\includegraphics[width=0.19\linewidth,height=2.9cm]{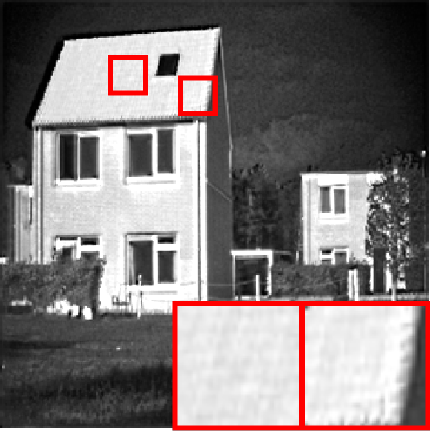} \\
		\small {(f) } \ & \small {(g) } \ & \small {(h) } \ & \small {(i) }\ & \small {(j) }\\
	\end{tabular}
    \end{center}
    \vspace{-1mm}
	\caption{Fusion result comparison of thermal radiation and visible texture information retention in infrared image and visible image, respectively. (a-j) correspond to the original visible and infrared images, and the fused results of GTF, DTCWT, DeepFuse, DenseFuse, IFCNN, U2Fusion, FusionGAN and our model.}
	\label{fig4}
\end{figure*}

\section{Experiments}
\subsection{Implementation}
The model is implemented with TensorFlow on GTX 2080TI GPU. Adam \cite{Adam} optimizer with the learning rate of 1e-4 is adopted. The batch size is 1, and the momentum value is 0.9. The weight decay is 5e-3. $\tau$ and $\xi$ are taken to 1.0 and 1.7, respectively. We train our model on the TNO database with 110 groups of infrared and visible images \footnote{https://figshare.com/articles/TN\_Image\_Fusion\_Dataset/1008029.}.

\subsection{Evaluation Criteria}\label{4.1}
We adopt six widely used objective metrics for evaluating the performance of our method and the competitors, e.g., Entropy (EN), Average gradient (AG), Gray level difference (GLD)~\cite{GLD}, Mutual information (MI)~\cite{MI}, Spatial frequency (SF)~\cite{SF} and visual information fidelity for fusion(VIFF)~\cite{VIF}, which are consistent with subjective visual evaluation.

\textbf{Average gradient (AG).}
AG reflects the clarity of fusion image, which represents the contrast of small details and local texture changes in the image. The larger AG, the more structure information retained in the fusion result.

\begin{equation}
\begin{aligned}
AG&=\frac {1}{(M-1)(N-1)}\sum_{i=1}^{M-1}\sum_{j=1}^{N-1}\\
&\sqrt{\frac{(I_f(i+1,j)-I_f(i,j))^{2}+(I_f(i,j+1)-I_f(i,j))^{2}}{2}},
\end{aligned}
\end{equation}
where $M$ and $N$ are the width and height of fusion image $I_f$, and (i,j) denotes the location of each pixel.

\textbf{Entropy (EN).}
Information theory-based EN denotes the average amount of information contained in an image, which is defined as:
\begin{equation}
En = - \sum_{i=1}^{L-1}p_ilog_2p_i,
\end{equation}
where $L$ denotes the gray scale of image, $p_i$ is the probability of gray  value $i$ appearing in the image. Larger $En$ represents richer information in the image.

\textbf{Mutual information (MI).}
MI can be used to measure the correlation between fused image and source images in the fusion field. Larger $MI$ represents more vital information contained in the fusion result.
\begin{equation}
MI=MI(I_1,I_f)+MI(I_2,I_f),
\end{equation}
where $MI(I_1,I_f)$ and $MI(I_2,I_f)$ denote the correlation among fusion result, infrared image, and visible image, respectively.  $MI(I_n,I_f)$ is defined as:
\begin{equation}
MI(I_n,I_f)=E(I_n)+E(I_f)-E(I_n,I_f),
\end{equation}
where $E(I_n)$ and $E(I_f)$ denote the information entropy of image $I_n$ and $I_f$, respectively. $E(I_n,I_f)$ is the joint information entropy.

\textbf{Gray level difference (GLD).}
GLD denotes the amount of the gradient information in the fused image. A larger GLD stands for more texture information contained in the fused result.
\begin{equation}
\begin{aligned}
GLD&=\frac {1}{(M-1)(N-1)}\sum_{i=1}^{M-1}\sum_{j=1}^{N-1}\mid(I_f(i,j)-I_f(i+1,j))\mid\\
&+\mid(I_f(i,j)-I_f(i,j+1))\mid,
\end{aligned}
\end{equation}

\textbf{Spatial frequency (SF).}
SF reflects the change of image gray level, which is consisted by horizontal and vertical gradients. Larger SF denotes clearer fusion result.
\begin{equation}
SF=\sqrt{H^{2}+V^{2}},
\end{equation}
where $H$ and $V$ are:
\begin{equation}
\begin{aligned}
H=\sqrt{\frac {1}{MN}\sum_{i=1}^{M}\sum_{j=2}^{N}\mid I_f(i,j)-I_f(i,j-1)\mid^{2}},
\end{aligned}
\end{equation}

\begin{equation}
\begin{aligned}
V=\sqrt{\frac {1}{MN}\sum_{i=2}^{M}\sum_{j=1}^{N}\mid I_f(i,j)-I_f(i-1,j)\mid^{2}}.
\end{aligned}
\end{equation}
 \begin{figure*}
    \begin{center}
	\begin{tabular}{c@{}c@{}c@{}c@{}c}
		\includegraphics[width=0.19\linewidth,height=2.9cm]{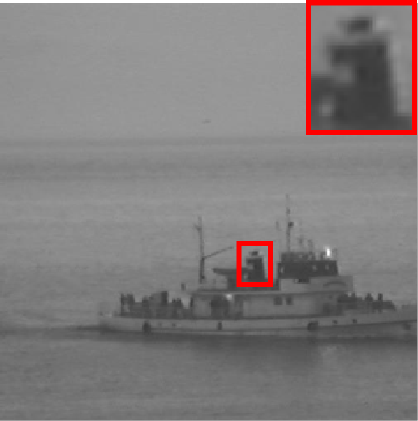} \ &
		\includegraphics[width=0.19\linewidth,height=2.9cm]{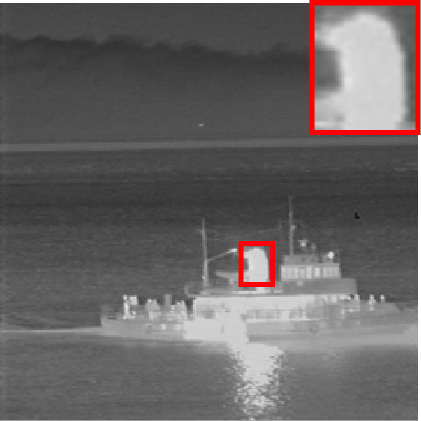} \ &
		\includegraphics[width=0.19\linewidth,height=2.9cm]{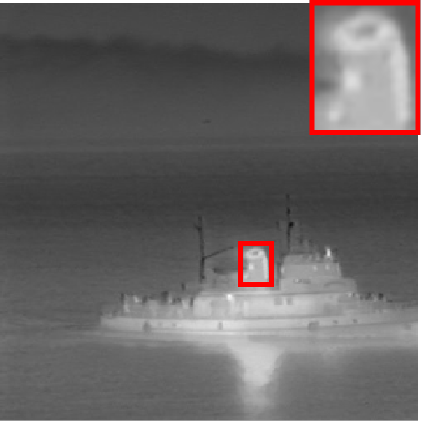} \ &
		\includegraphics[width=0.19\linewidth,height=2.9cm]{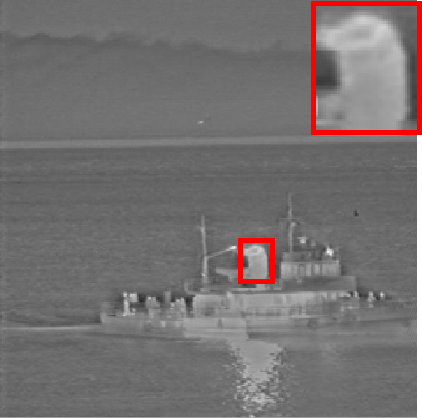}  \ &
        \includegraphics[width=0.19\linewidth,height=2.9cm]{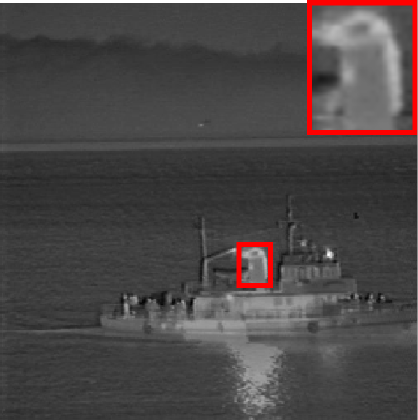} \\
		\small {(a1) } \ & \small {(a2) } \ & \small {(a3) } \ & \small {(a4) }\ & \small {(a5) }\\		
		\includegraphics[width=0.19\linewidth,height=2.9cm]{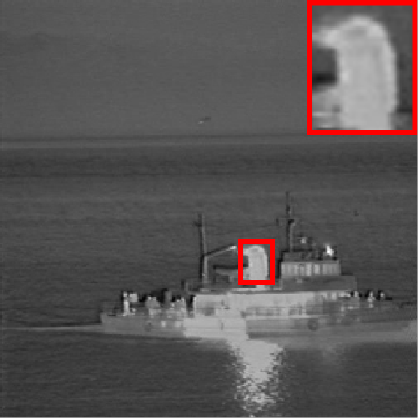} \ &
		\includegraphics[width=0.19\linewidth,height=2.9cm]{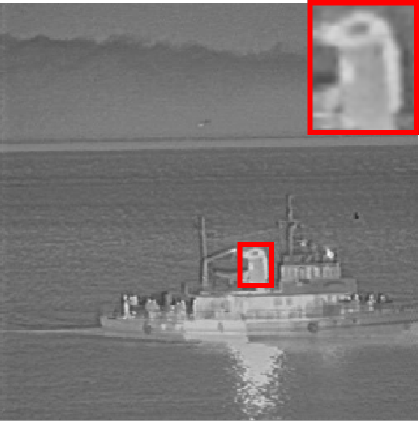} \ &
        \includegraphics[width=0.19\linewidth,height=2.9cm]{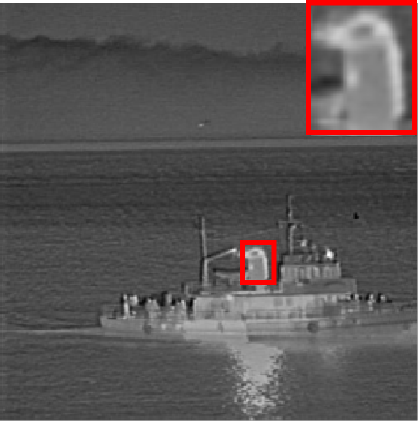} \ &
		\includegraphics[width=0.19\linewidth,height=2.9cm]{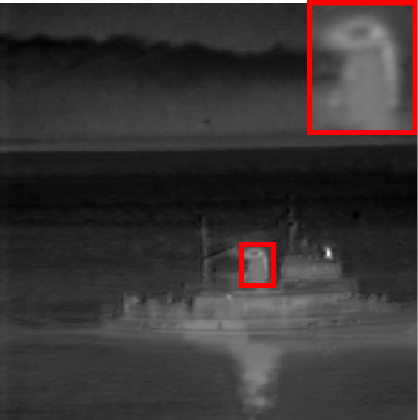}\ &
		\includegraphics[width=0.19\linewidth,height=2.9cm]{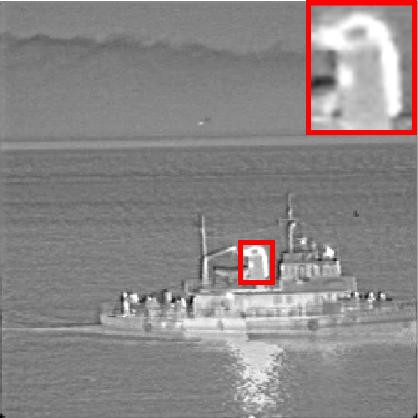} \\
		\small {(a6) } \ & \small {(a7) } \ & \small {(a8) } \ & \small {(a9) }\ & \small {(a10) }\\
				\includegraphics[width=0.19\linewidth,height=2.9cm]{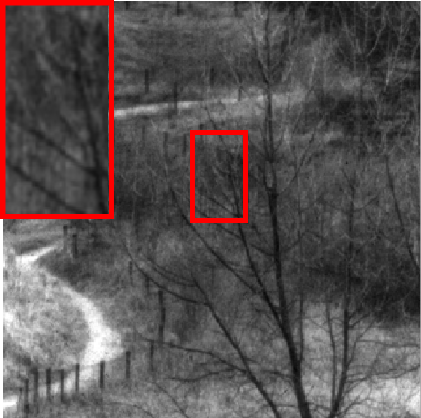} \ &
		\includegraphics[width=0.19\linewidth,height=2.9cm]{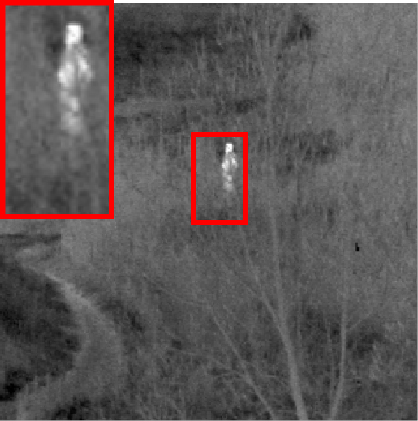} \ &
		\includegraphics[width=0.19\linewidth,height=2.9cm]{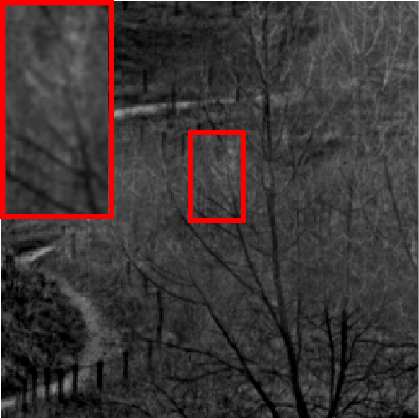} \ &
		\includegraphics[width=0.19\linewidth,height=2.9cm]{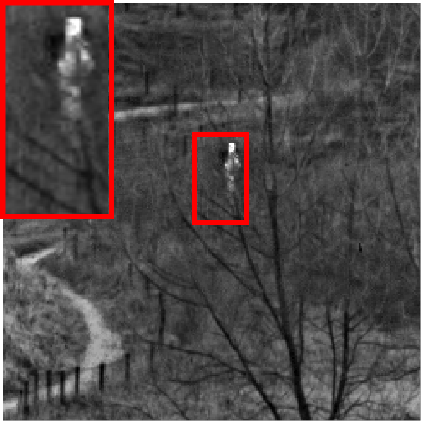}  \ &
        \includegraphics[width=0.19\linewidth,height=2.9cm]{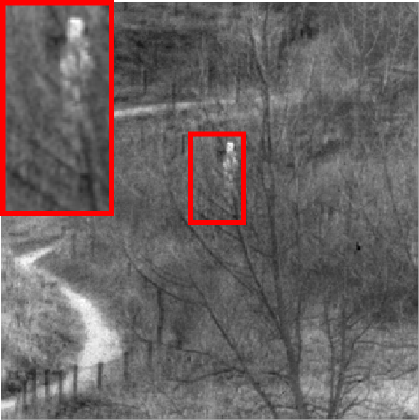} \\
		\small {(b1) } \ & \small {(b2) } \ & \small {(b3) } \ & \small {(b4) }\ & \small {(b5) }\\	
		\includegraphics[width=0.19\linewidth,height=2.9cm]{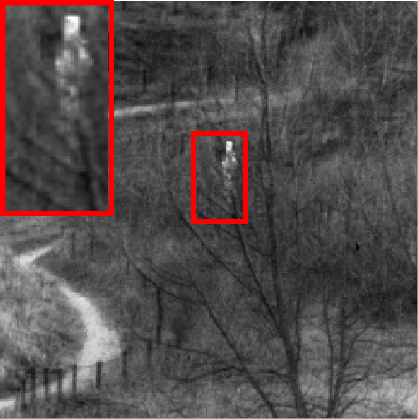} \ &
		\includegraphics[width=0.19\linewidth,height=2.9cm]{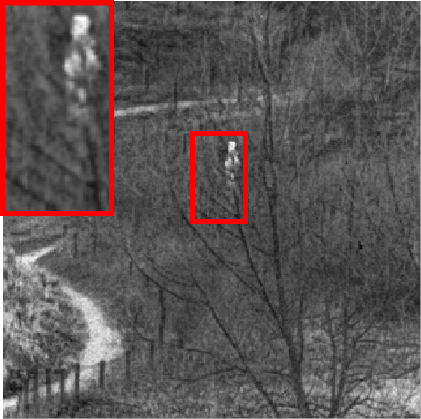} \ &
        \includegraphics[width=0.19\linewidth,height=2.9cm]{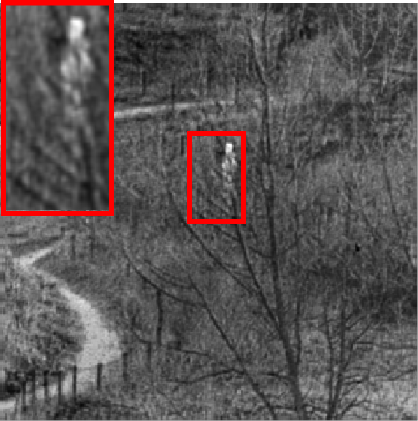} \ &
		\includegraphics[width=0.19\linewidth,height=2.9cm]{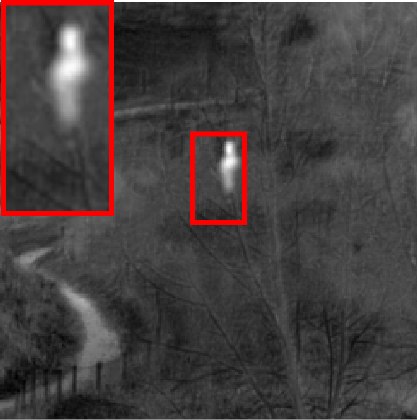}\ &
		\includegraphics[width=0.19\linewidth,height=2.9cm]{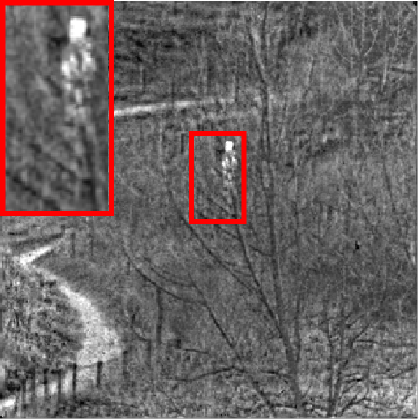} \\
		\small {(b6) } \ & \small {(b7) } \ & \small {(b8) } \ & \small {(b9) }\ & \small {(b10) }\\
	\end{tabular}
    \end{center}
    \vspace{-1mm}
	\caption{Fusion result comparison of thermal radiation and visible texture information retention in infrared image and visible image, respectively. (a1-a10) and (b1-b10) correspond to the original visible and infrared images, and the fused results of GTF, DTCWT, DeepFuse, DenseFuse, IFCNN, U2Fusion, FusionGAN and our model.}
	\label{fig5}
\end{figure*}
\textbf{Visual information fidelity for fusion (VIFF).}
Since we aim to mitigate the vital information loss in the fused result, we adopt another evaluation matrix, named as VIFF~\cite{VIF}. VIFF measures the amount of visible information retained in the fused image, which is consistent with the human visual system. Thus, a larger VIFF represents that more visible information are fused into the fusion image, and less distortion between the fused result and source images.

\subsection{Results and Analysis}\label{4.2}

In this section, we conduct qualitative and quantitative evaluations. The experiments are conducted on two widely used datasets for infrared and visible image fusion, named as INFV-20 dataset \footnote{https://github.com/hli1221/imagefusion\_densefuse.} and INFV-41 dataset \footnote{https://github.com/jiayi-ma/FusionGAN.}, respectively. For each dataset, we compare our model with seven state-of-the-art methods, including two hand-crafted feature-based methods, four non-adversarial CNN-based fusion methods, and one adversarial CNN-based fusion method, i.e., gradient transfer fusion (GTF)~\cite{Ma2016Infrared}, DTCWT~\cite{li2011performance}, DeepFuse~\cite{deepfuse}, DenseFuse~\cite{Li2018}, IFCNN~\cite{Zhang2019}, U2Fusion~\cite{9151265}, and FusionGAN~\cite{Ma2019s}.

\subsubsection{Quantitative Evaluation}\label{4.2.1}

We firstly conduct quantitative comparison between our results and the results generated by the competitors using AG, EN, MI, GLD, SF and VIFF metrics on INFV-20 and INFV-41. Table I summarizes the average quality metrics of different methods on INFV-41 dataset. Obviously, our method achieves the best performance in terms of AG, EN, MI, GLD and SF metrics, and subprime performance on VIFF metric. The largest values of EN, AG, GLD and SF indicate that larger gradient, richer texture and higher contrast information are retained in the results. In addition, the satisfied values of MI and VIFF denote higher similarity between the result and source image, and meanwhile more visible information are retained in fusion image. Precisely, this proves the starting point of our approach that aims to fuse more vital information. Same conclusion can be drawn from Table II.
Especially, DenseFuse and U2Fusion focus on extracting multi-level features contained in both visible and thermal images.
Thus, they achieve better results of VIFF and MI, since the fused image retain the universal features of source images.
In contrast, our fused image not only retains the universal features of source images, but also fuses discrepancy features of source images, thereby obtaining better results of AG, GLD and SF (where they comprehensively evaluate the performance of fused results, such as universal feature and discrepancy feature retention).

\subsubsection{Qualitative Evaluation}
\begin{figure*}
    \begin{center}
	\begin{tabular}{c@{}c@{}c@{}c@{}c}
		\includegraphics[width=0.19\linewidth,height=2.9cm]{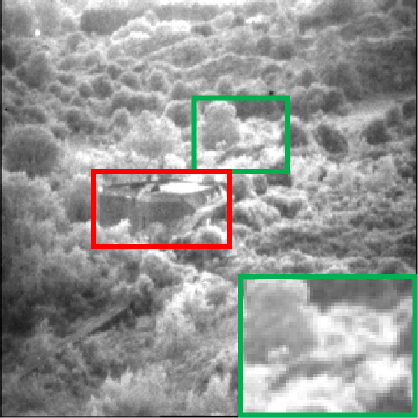} \ &
		\includegraphics[width=0.19\linewidth,height=2.9cm]{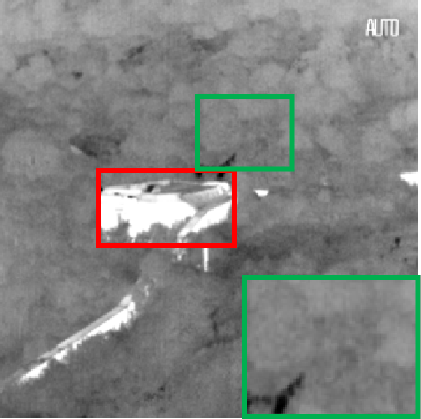} \ &
		\includegraphics[width=0.19\linewidth,height=2.9cm]{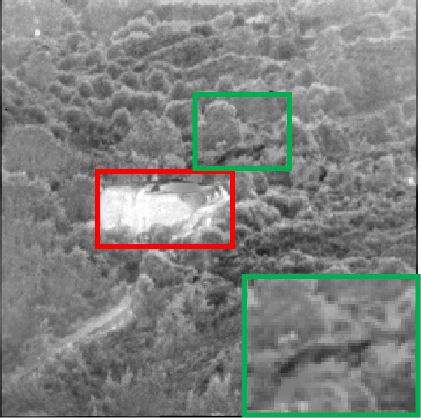} \ &
		\includegraphics[width=0.19\linewidth,height=2.9cm]{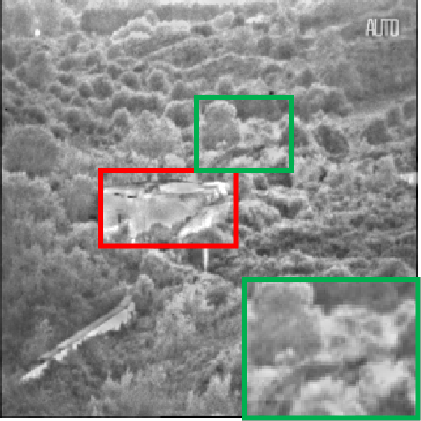}  \ &
		\includegraphics[width=0.19\linewidth,height=2.9cm]{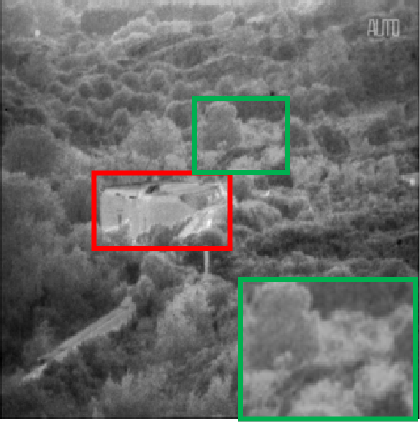} \\
		\small {(a) } \ & \small {(b) } \ & \small {(c) } \ & \small {(d) }\ & \small {(e) }\\
		\includegraphics[width=0.19\linewidth,height=2.9cm]{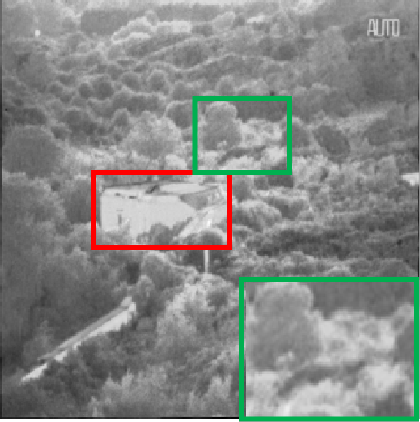} \ &
		\includegraphics[width=0.19\linewidth,height=2.9cm]{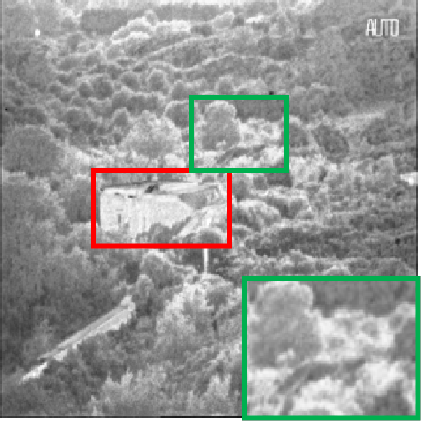} \ &
        \includegraphics[width=0.19\linewidth,height=2.9cm]{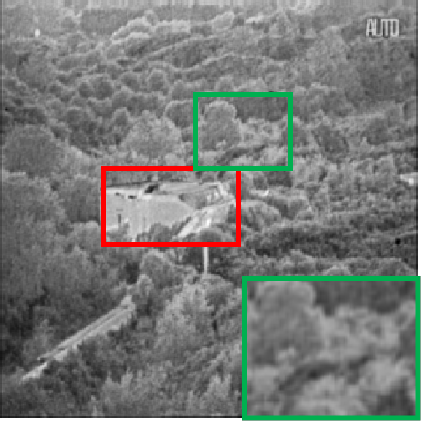} \ &
		\includegraphics[width=0.19\linewidth,height=2.9cm]{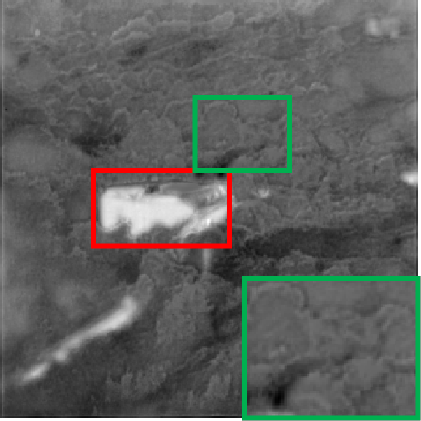}\ &
		\includegraphics[width=0.19\linewidth,height=2.9cm]{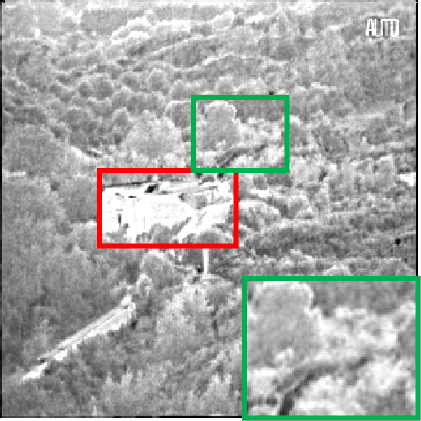} \\
		\small {(f) } \ & \small {(g) } \ & \small {(h) } \ & \small {(i) }\ & \small {(j) }\\
	\end{tabular}
    \end{center}
    \vspace{-1mm}
	\caption{Fusion result comparison of vital information retention in the visible image. (a-j) correspond to the original visible and infrared images, and the fused results of GTF, DTCWT, DeepFuse, DenseFuse, IFCNN, U2Fusion, FusionGAN and our model.}
	\label{fig6}
\end{figure*}

\begin{figure*}
    \begin{center}
	\begin{tabular}{c@{}c@{}c@{}c@{}c}
		\includegraphics[width=0.19\linewidth,height=2.9cm]{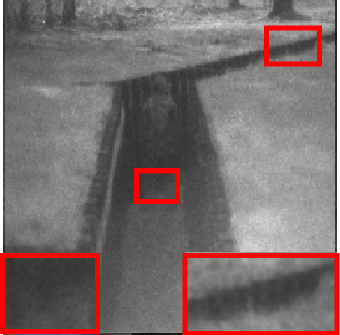} \ &
		\includegraphics[width=0.19\linewidth,height=2.9cm]{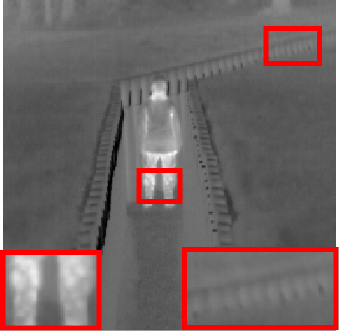} \ &
		\includegraphics[width=0.19\linewidth,height=2.9cm]{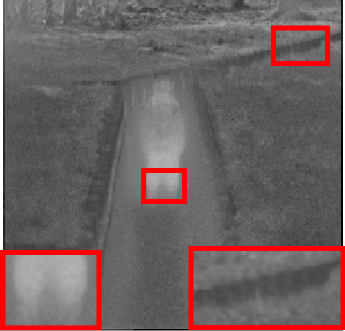} \ &
		\includegraphics[width=0.19\linewidth,height=2.9cm]{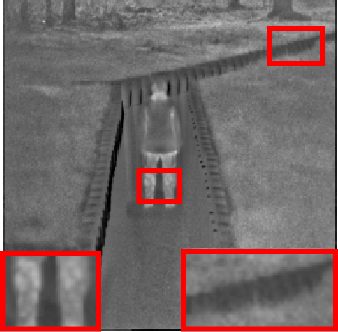}  \ &
		\includegraphics[width=0.19\linewidth,height=2.9cm]{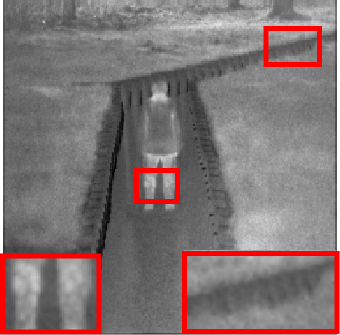} \\
		\small {(a) } \ & \small {(b) } \ & \small {(c) } \ & \small {(d) }\ & \small {(e) }\\
		\includegraphics[width=0.19\linewidth,height=2.9cm]{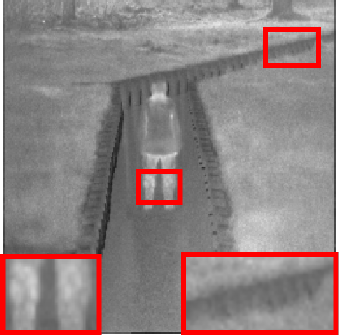} \ &
		\includegraphics[width=0.19\linewidth,height=2.9cm]{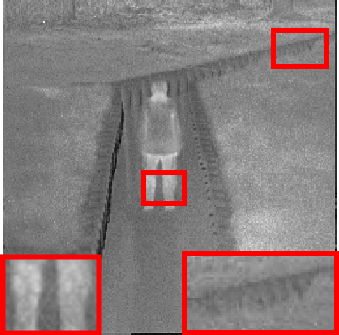} \ &
        \includegraphics[width=0.19\linewidth,height=2.9cm]{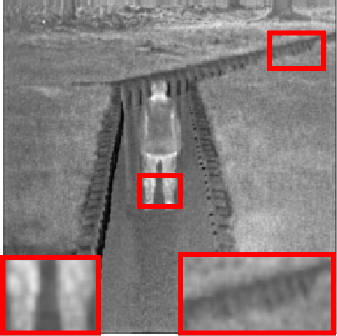} \ &
		\includegraphics[width=0.19\linewidth,height=2.9cm]{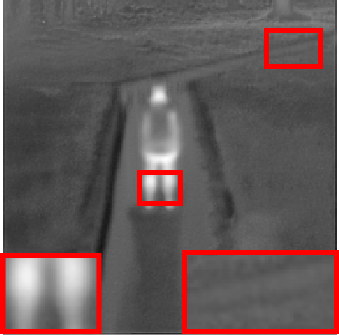}\ &
		\includegraphics[width=0.19\linewidth,height=2.9cm]{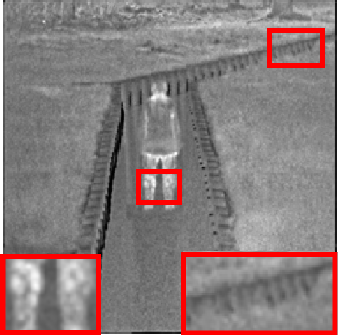} \\
		\small {(f) } \ & \small {(g) } \ & \small {(h) } \ & \small {(i) }\ & \small {(j) }\\
	\end{tabular}
    \end{center}
    \vspace{-1mm}
	\caption{Fusion result comparison of vital information retention in the infrared image. (a-j) correspond to the original visible and infrared images, and the fused results of GTF, DTCWT, DeepFuse, DenseFuse, IFCNN, U2Fusion, FusionGAN and our model.}
	\label{fig7}
\end{figure*}

We then conduct qualitative comparison with competitors. Visual comparisons on five pairs of representative images are provided in Figs.~\ref{fig4}-\ref{fig7}. In the subsection, we compare the retention of vital features of source images including both universal features and domain discrepancy features.

\begin{figure}
    \begin{center}
	\begin{tabular}{c@{}c@{}c}
		\includegraphics[width=0.3\linewidth,height=2.3cm]{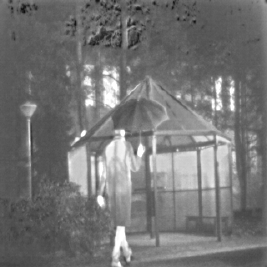} \ &
		\includegraphics[width=0.3\linewidth,height=2.3cm]{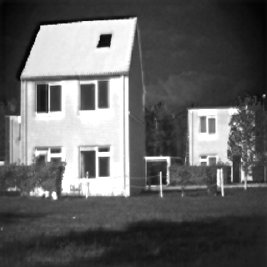} \ &
		\includegraphics[width=0.3\linewidth,height=2.3cm]{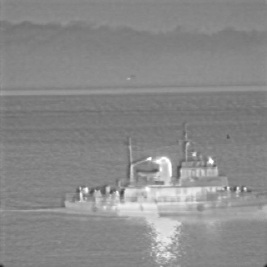}   \\
			\includegraphics[width=0.3\linewidth,height=2.3cm]{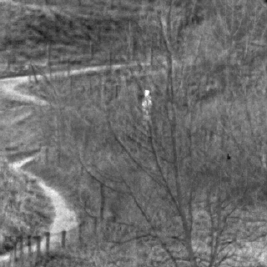} \ &
		\includegraphics[width=0.3\linewidth,height=2.3cm]{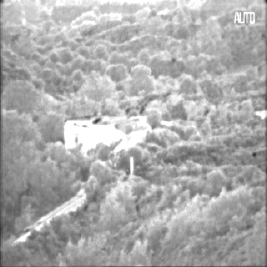} \ &
		\includegraphics[width=0.3\linewidth,height=2.3cm]{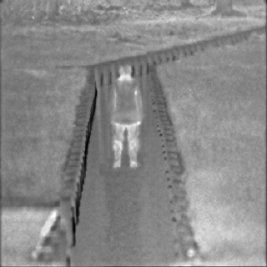}   \\
	\end{tabular}
    \end{center}
    \vspace{-1mm}
	\caption{Fusion results by our method with the smooth constraint. Noise is suppressed in the fused images. The sources are in Fig. 1 and Figs. 4-7.}
	\label{fig8a}
\end{figure}
\textbf{Retaining of thermal radiation and texture information.} As shown in Fig.~\ref{fig4}(a) and Fig.~\ref{fig4}(b), visible image is represented as gradient details while infrared image is mainly characterized as thermal radiation information. Thus, we evaluate the fusion performance from the perspective of the retention degree of such information.

As illustrated in Fig.~\ref{fig4}, on the whole, all methods can fuse the main features of source images to some extent. More concretely, GTF can greatly fuse the thermal radiation information with high pixel intensity, while the structure features in visible image are lost. This can be explained that GTF aims to preserve the main intensity information of infrared image and gradient variation of visible image. As shown in the rectangular box of Fig.~\ref{fig4}(c), the fusion result
has strong intensity similarity with the infrared image, but the visible information (e.g., the roof of tile) is lost. As shown in the rectangular box of Fig.~\ref{fig4}(d-e), both NSCT-SR and DTCWT cannot well retain the high intensity thermal information of infrared image, and loss partial of texture information of visible image. Overall, those methods
based on hand-crafted features can not well handle domain discrepancy between source images. Since low-level features cannot sufficiently represent thermal radiation information and visible appearance information, resulting in vital features missing.

Fig.~\ref{fig4}(f-h) shows fusion results generated from non-adversarial based fusion methods. Intuitively, DeepFuse, DenseFuse and IFCNN cannot highlight the thermal information well, and partial gradient details of visible image are lost. We explain that those fusion methods focus on extracting and preserving vital features utilizing CNN structures via unsupervised strategy. On one hand, utilizing same convolution operator for generating weight maps or feature extraction, consequently leads to the loss of vital features which are specific to source images with domain discrepancy. On the other hand, the unsupervised training is achieved by optimizing loss function to constrain the fusion results containing the main contents of source images. Similarly, as shown in Fig.~\ref{fig4}(i), FusionGAN based on adversarial training, also loses vital information, accompanied by ambiguity and distortion. Thus, the unsupervised mechanism is unable to extract features adequately, which subsequently cannot guarantee all vital information of source images can be retained. In contrast, as shown in Fig.~\ref{fig4}(j), the vital information, including thermal information of infrared image and gradient details of visible image, are well retained in our fused result. Fig.~\ref{fig5} presents the fusion results on two representative image pairs. Obviously, our results also appear better fusion performance containing more vital information.
\begin{figure*}
    \begin{center}
	\begin{tabular}{c@{}c@{}c}
		\includegraphics[width=0.20\linewidth,height=2.6cm]{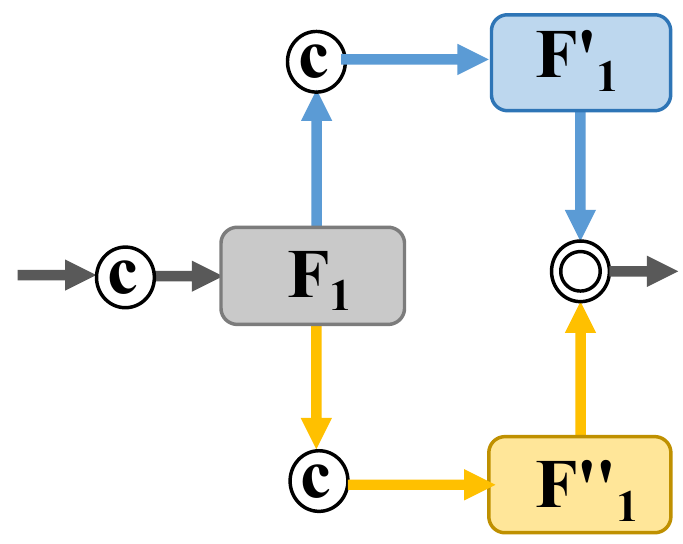} \ &
        \includegraphics[width=0.10\linewidth,height=2.6cm]{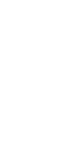} \ &
		\includegraphics[width=0.58\linewidth,height=2.6cm]{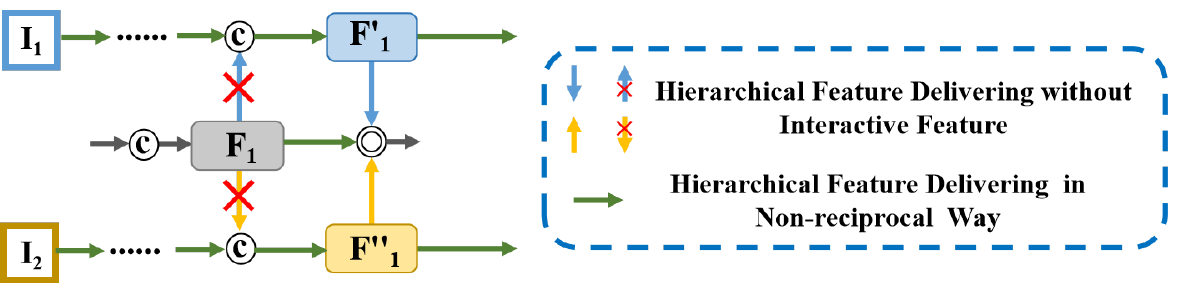}  \\
		\small {(a) } \ & \small { }\ & \small {(b) }\\
	\end{tabular}
    \end{center}
    \vspace{-1mm}
	\caption{(a) Basic constituent block (single stage IFEM) of IFESNet. (b) Basic constituent block of IFESNet w/o IFEM. IFESNet w/o IFEM adopts the same structure with IFESNet, but different feature stream. Instead of utilizing interactive feature between IVIFNet and SHFENet for gradually increasing extra vital information extraction, IFESNet w/o IFEM conducts hierarchical feature delivering in non-reciprocal way. }
	\label{fig8}
\end{figure*}

\textbf{Retaining of other vital features.}
As discussed previously, not only thermal information with high pixel intensity is presented in infrared image, whereas, the other characteristics, such as gradient variation, texture and edge information are also included. In the same way, visible images also contain the other vital features, such as intensity, contrast and saturability. In this section,
we further evaluate the retaining of other vital features of source images.

Fig.~\ref{fig6} presents the comparisons. It's worth noting that, here, we mainly focus on the retention of vital information in the visible image, in spite that our results can also maximum fusion of thermal radiation information in infrared images with higher contrast (as shown in the red rectangle box). More concretely, Fig.~\ref{fig6}(a) presents abundant appearance,
including high intensity, contrast and saturability information (as shown in the green rectangle box). As illustrated in Fig.~\ref{fig6}(c), GTF can preserve the gradient variations of visible image, whereas the other vital information such as intensity, contrast and saturability is lost. The reason is that these features are not taken into account when modeling. The same issue appears in Fig.~\ref{fig6}(d-e), which are generated by NSCT-SR and DTCWT, respectively. IFCNN can retain much more vital information from visible image compared with DeepFuse and DenseFuse. However, compared with visible image and our result,
IFCNN still encounters certain degree of loss in brightness, contrast and saturation.

In addition, we further evaluate the retention of other vital features from infrared image, such as edge, gradient and texture information. A visual comparison is provided in Fig.~\ref{fig7}. As shown in the highlighted regions of Fig.~\ref{fig7}(b), the infrared image presents some texture information which is almost invisible in Fig.~\ref{fig7}(a). Overall, as shown in the red boxes of Fig.~\ref{fig7}(j), our result exhibits more texture appearance than the other methods. As illustrated in Fig.~\ref{fig7}(c-i), the competitors lose some details, e.g., texture of legs, and edge of wall. This limits the application in scenario where gradient information is unavailable in visible image, whereas abundance in infrared image. Thus, our method can also preserve the detail information of infrared image more completely, which is ignored by other methods since they mainly focus on the thermal radiation information.
Nevertheless, we attribute the excellent performance of our method to: 1) self-supervised strategy can generate more and comprehensive features of source image; 2)  multi-stage interactive feature embedding learning can gradually integrate all vital information into fusion results and thus solve the vital information missing problem.

\textbf{Discussion.}
Since image fusion focuses on generating a new image that retains source images' details, fused image will contain noise if the source image (e.g., infrared image) contains noise, as shown in Figs. 4-7.
Here, we add a smooth constraint for the weight maps (see Eq. (4)) by using a Gaussian filter with variance 2 and window size $5\times 5$. The noise problem is relieved, as shown in Fig.~\ref{fig8a}. More denoising strategies (e.g., feature denoising) will be studied in future work.

\subsection{Ablation Study}\label{4.4}
\textbf{Effect of Interactive Feature Embedding Learning.}
 As described in Section III-B, interactive feature embedding model (IFEM) is designed, for promoting hierarchical feature extraction and fusion in the bi-directional interactive way. To analyze the contribution of this mechanism, we implement a variant named IFESNet w/o IFEM for comparison. The basic unit of IFESNet and IFESNet w/o IFEM are provided in Fig.~\ref{fig8}(a) and Fig.~\ref{fig8}(b). As shown in Fig.~\ref{fig8}(b), the stage of IFESNet w/o IFEM is designed with the same architecture compared with Fig.~\ref{fig8}(a), but the data flow direction is different. To be specific, instead of using interactive feature embedding learning mechanism between fusion and reconstruction tasks in IFESNet, IFESNet w/o IFEM
only adopts hierarchical feature delivering from two reconstruction network without feature reverse delivering process from fusion network. Hence, hierarchical feature delivering is conducted in a non-reciprocal way. For a fair comparison,
 IFESNet and IFESNet w/o IFEM adopt identical convolution layers with the same parameters. Quantitative evaluations are shown in Table III and Table IV. Compared with IFESNet, IFESNet w/o IFEM results in poor fusion performance with all indexes decreased significantly. Although IFESNet w/o IFEM conducts hierarchical feature extraction in the self-supervised way, those hierarchical features without stage-interactive can't guarantee to contain vital information sufficiently for fusion.

\begin{table}
	\small
	\renewcommand{\arraystretch}{1.5}
	\begin{center}
		\caption{Effect of hierarchical feature extraction via self-supervised strategy on INFV-41 dataset.}
		\label{tab:precision1}
		\vspace{0.1cm}
		\begin{tabular}{cp{0.5cm}p{0.5cm}p{0.5cm}p{0.5cm}p{0.5cm}p{0.5cm}}
			\toprule[1pt]
			& AG & EN & MI & GLD & SF & VIFF\\
			\hline\hline
			IFESNet w/o IFEM  &{9.420}  &{6.879} &{13.76} &{16.57} &{.0741}  &{.5775}\\	
			\cline{1-7}
			IFESNet  &\textbf{{10.18}}  &\textbf{{6.923}} &\textbf{{13.85}} &\textbf{{17.96}} &\textbf{{.0787}} &\textbf{{.6414}} \\		
			\toprule[1pt]
		\end{tabular}
	\end{center}
	\vspace{-1mm}
\end{table}

\begin{table}
	\small
	\renewcommand{\arraystretch}{1.5}
	\begin{center}
		\caption{Effect of hierarchical feature extraction via self-supervised strategy on INFV-20 dataset.}
		\label{tab:precision1}
		\vspace{0.1cm}
		\begin{tabular}{cp{0.5cm}p{0.5cm}p{0.5cm}p{0.5cm}p{0.5cm}p{0.5cm}}
			\toprule[1pt]
			& AG & EN & MI & GLD & SF & VIFF\\
			\hline\hline
			IFESNet w/o IFEM  &{9.102}  &{6.719}  &{13.44} &{15.53} &{.0720}  &{.6161}\\	
			\cline{1-7}
			IFESNet  &\textbf{{9.924}}  &\textbf{{6.809}} &\textbf{{13.62}} &\textbf{{16.99}} &\textbf{{.0770}} &\textbf{{.7011}} \\		
			\toprule[1pt]
		\end{tabular}
	\end{center}
	\vspace{-1mm}
\end{table}

\begin{figure}
    \begin{center}
	\begin{tabular}{c}
		\includegraphics[width=0.9\linewidth,height=4.0cm]{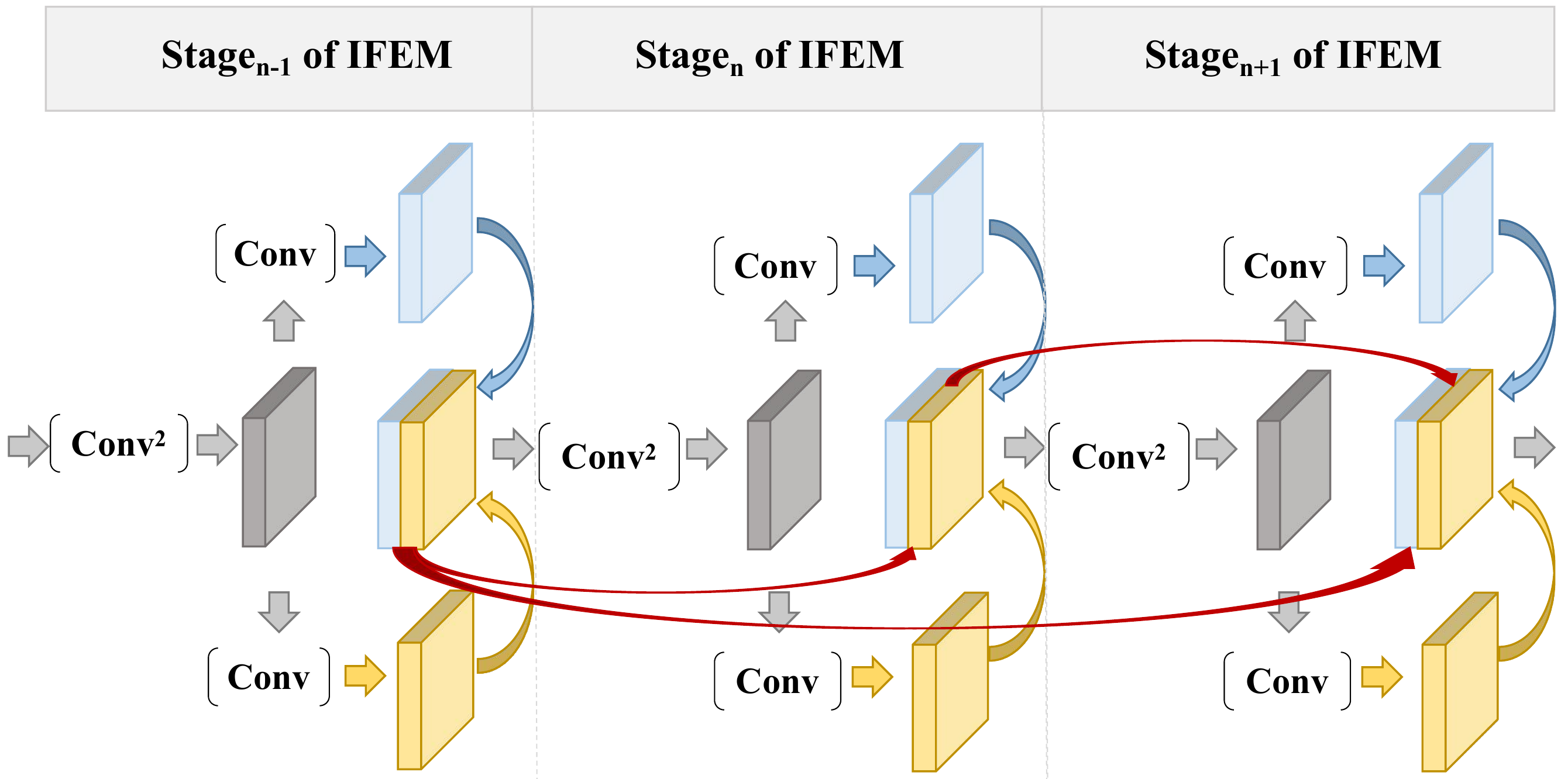}  \\
	\end{tabular}
    \end{center}
    \vspace{-1mm}
	\caption{Illustration of hierarchical connection of each stage. The feature generated by the $n-1$th stage is skipped (red arrows) to its later stage's features (e.g., the $n$th stage and the $n+1$th stage) by concatenation operation.}
	\label{aa}
\end{figure}

\begin{table}[t]
	\small
	\renewcommand{\arraystretch}{1.5}
	\begin{center}
		\caption{Effect of varied blocks for fusion performance on INFV-20 dataset.}
		\label{tab:precision1}
		\vspace{0.1cm}
		\begin{tabular}{cp{0.5cm}p{0.5cm}p{0.5cm}p{0.5cm}p{0.5cm}p{0.5cm}}
			\toprule[1pt]
			& AG & EN & MI & GLD & SF & VIFF\\
			\hline\hline
			IFESNet-S1  &{{10.20}}  &{6.758} &{13.52} &{{17.52}} &{{.0796}}  &{.6705}\\	
			\cline{1-7}
			IFESNet-S2  &{9.821}  &{6.756} &{13.52} &{16.76} &{.0777} &{.6751} \\	
			\cline{1-7}
			IFESNet-S3  &{9.924}  &{{6.809}} &{{13.62}} &{16.99} &{.0770} &{.7011} \\	
			\cline{1-7}
			IFESNet-S4 &{9.670}  &{6.787} &{13.58} &{16.49} &{.0767} &{{.7209}}\\
            \cline{1-7}\hline\hline
			IFESNet-HC &\textbf{11.10}  &{\textbf{6.817}} &\textbf{{13.64}} &\textbf{18.94} &\textbf{.0860} &\textbf{{.7566}}\\		
			\toprule[1pt]
		\end{tabular}
	\end{center}
	\vspace{-1mm}
\end{table}

\textbf{Effect of Varied IFEM Stages.} As shown in Fig.~\ref{fig8}(a), we regard each layer of IFESNet and its corresponding layer of IVIFNet with one interactive feature embedding learning process, as a IFEM stage. In this study, we totally adopt three IFEM stages for vital feature extraction and fusion. In this section, we aim to analyze the performance of IFESNet with varied IFEM stages. To be specific, we compare IFESNet with one-stage (named as IFESNet-S1), two-stages (named as IFESNet-S2), three-stages (named as IFESNet-S3) and four-stages (named as IFESNet-S4). Quantitative evaluation of IFESNet and three variants on INFV-20 dataset and INFV-41 dataset are presented in Table V and Table VI, respectively.

As illustrated in Table V and Table VI, with the increase of stages, the evaluation criteria based on gradient information (e.g., AG, GLD and SF) tend to be smaller, while the evaluation criteria based on information entropy (e.g., EN), image similarity (e.g., MI) and fidelity (e.g., VIFF) tend to be larger. This can be explained as follows: when IFESNet adopts only a stage, the network itself pays more attention to the low-level features, such as texture and edge information. Thus the indexes based on gradient are higher. However, with the stage number increases, more high-level semantic information is extracted and then fused.
In this case, the fusion appearance is closer to source image with more vital features while achieving better fidelity. Thus, by judging and weighing the fusion of both high-level and low-level features, three stages are adopted in the experiment.

As described above, with deepening of the network, more high-level semantic information is extracted and fused. Thus, the performance of the pixel-level fusion task may be limited.
The hierarchical connection of each stage (denoted as IFESNet-HC) can deal with this dilemma, as shown in Fig.~\ref{aa}.
The low-level and high-level features are integrated, thereby improving the fusion performance.
In Table V, the hierarchical connection of each stage makes the deep network (four stages) achieve the best results.

\textbf{Effect of self-supervised reconstruction loss.} As described in Section III-D, we adopt MSE as loss function for self-supervised hierarchical feature extraction network training.
Here, the MAE-based perceptual loss in the self-supervised mechanism is used for comparison. As shown in Table VI, our MSE-based loss can
achieve better fusion performance.

\begin{table}[t]
	\small
	\renewcommand{\arraystretch}{1.5}
	\begin{center}
		\caption{Effect of self-supervised reconstruction loss for fusion performance on INFV-20 dataset.}
		\label{tab:precision1}
		\vspace{0.1cm}
		\begin{tabular}{cp{0.5cm}p{0.5cm}p{0.5cm}p{0.5cm}p{0.5cm}p{0.5cm}}
			\toprule[1pt]
			& AG & EN & MI & GLD & SF & VIFF\\
			\hline\hline
			MAE-based loss &{9.762}  &{6.773} &{13.55} &{16.77} &{.0765}  &{.6997}\\	
			\cline{1-7}
			MSE-based loss &\textbf{{9.924}}  &\textbf{{6.809}} &\textbf{{13.62}} & \textbf{{16.99}} &\textbf{{.0770}} &\textbf{{.7011}} \\		
			\toprule[1pt]
		\end{tabular}
	\end{center}
	\vspace{-1mm}
\end{table}

\begin{table}[t]
	\small
	\renewcommand{\arraystretch}{1.5}
	\begin{center}
		\caption{Effect of structural similarity index loss for fusion performance on INFV-20 dataset.}
		\label{tab:precision1}
		\vspace{0.1cm}
		\begin{tabular}{cp{0.5cm}p{0.5cm}p{0.5cm}p{0.5cm}p{0.5cm}p{0.5cm}}
			\toprule[1pt]
			& AG & EN & MI & GLD & SF & VIFF\\
			\hline\hline
			VP-based loss &{8.417}  &{6.451} &{12.90} &{14.12} &{.0639}  &{.2425}\\	
			\cline{1-7}
			SSIM-based loss &\textbf{{9.924}}  &\textbf{{6.809}} &\textbf{{13.62}} & \textbf{{16.99}} &\textbf{{.0770}} &\textbf{{.7011}} \\		
			\toprule[1pt]
		\end{tabular}
	\end{center}
	\vspace{-1mm}
\end{table}

\textbf{Effect of structural similarity index loss.}
In order to retain source image's structural details,
In Section III-D, we adopt structural similarity index metric (SSIM)-based loss to train IVIFNet.
Here, the visible perception loss (VP)~\cite{zhao2020learning} is adopted for comparison. As shown in Table VII, our SSIM-based loss obtain better fusion performance, since the fused image preserves source image's structural details.

\section{CONCLUSION}
In this paper, a novel interactive feature embedding in self-supervised learning  framework for infrared and visible image fusion is proposed for improving the vital information retention in fusion results. In particular, the self-supervised strategy is designed for capturing more informative representations of source images, which are adequately for jointly source image reconstruction and fusion.
Moreover, stage-interactive feature embedding learning mechanism between a fusion network and two reconstruction networks is designed for embedding the vital information through stage-wise hierarchical feature interaction, which essentially is implemented by leveraging
all the hierarchical features from different tasks. Qualitatively and quantitatively comparisons with the state of the arts indicate our method can not only better fuse the thermal radiation information of infrared image and the structural information of visible image, but also can retain the other vital information in infrared image (e.g., texture, edge) and visible image (e.g., intensity, contrast, saturation).

\bibliographystyle{IEEEtran}
\bibliography{IEEEabrv,IF_r3_A}

\end{document}